\documentclass[10pt,journal,compsoc]{IEEEtran}

\ifCLASSOPTIONcompsoc
  \usepackage[nocompress]{cite}
\else
  \usepackage{cite}
\fi
\usepackage{amsmath,amsfonts}
\usepackage{algorithmic}
\usepackage{algorithm}
\usepackage{array}
\usepackage[caption=false,font=normalsize,labelfont=sf,textfont=sf]{subfig}
\usepackage{textcomp}
\usepackage{stfloats}
\usepackage{url}
\usepackage{verbatim}
\usepackage{graphicx}
\usepackage{cite}
\usepackage{bm}
\usepackage{amssymb}
\usepackage{booktabs}
\usepackage{overpic}
\usepackage[pagebackref,breaklinks,colorlinks]{hyperref}
\usepackage{hyperref}

\ifCLASSINFOpdf
 
\else

\fi

\begin{document}
\title{Neural-ABC: Neural Parametric Models for Articulated Body with Clothes}

\author{Honghu~Chen,
        Yuxin~Yao,
        and~Juyong~Zhang$^\dagger$,~\IEEEmembership{Member,~IEEE,}
\IEEEcompsocitemizethanks{\IEEEcompsocthanksitem  H. Chen is with the School
        of Data Science, University of Science and Technology of China.}
\IEEEcompsocitemizethanks{\IEEEcompsocthanksitem  Y. Yao, and J. Zhang are with the School
of Mathematical Science, University of Science and Technology of China.
}

	\thanks{$^\dagger$Corresponding author. Email: \texttt{juyong@ustc.edu.cn}.}
}

\markboth{Submitted to IEEE TRANSACTIONS ON VISUALIZATION AND COMPUTER GRAPHICS}%
{Shell \MakeLowercase{\textit{et al.}}: Bare Demo of IEEEtran.cls for Computer Society Journals}

\IEEEtitleabstractindextext{
\begin{abstract}
  In this paper, we introduce Neural-ABC, a novel parametric model based on neural implicit functions that can represent clothed human bodies with disentangled latent spaces for identity, clothing, shape, and pose. 
  Traditional mesh-based representations struggle to represent articulated bodies with clothes due to the diversity of human body shapes and clothing styles, as well as the complexity of poses. 
  Our proposed model provides a unified framework for 
  parametric modeling, which can represent the identity, clothing, shape and pose of the clothed human body.
  Our proposed approach utilizes the power of neural implicit functions as the underlying representation and integrates well-designed structures to meet the necessary requirements.
  Specifically, we represent the underlying body as a signed distance function and clothing as an unsigned distance function, and they can be uniformly represented as unsigned distance fields.
  Different types of clothing do not require predefined topological structures or classifications, and can follow changes in the underlying body to fit the body.
  Additionally, we construct poses using a controllable articulated structure. 
  The model is trained on both open and newly constructed datasets, and our decoupling strategy is carefully designed to ensure optimal performance.
  Our model excels at disentangling clothing and identity in different shape and poses while preserving the style of the clothing. 
  We demonstrate that Neural-ABC fits new observations of different types of clothing. 
  Compared to other state-of-the-art parametric models, Neural-ABC demonstrates powerful advantages in the reconstruction of clothed human bodies,
  as evidenced by fitting raw scans, depth maps and images.
  We show that the attributes of the fitted results can be further edited by adjusting their identities, clothing, shape and pose codes. 
  The dataset and trained parametric model will be available at \href{https://ustc3dv.github.io/NeuralABC/}{https://ustc3dv.github.io/NeuralABC/}.
  
\end{abstract}

\begin{IEEEkeywords}
  neural implicit function, parametric human body model.
\end{IEEEkeywords}}

\maketitle

\IEEEdisplaynontitleabstractindextext

\begin{figure*}[ht]
  \centering
  \includegraphics[scale=0.26]{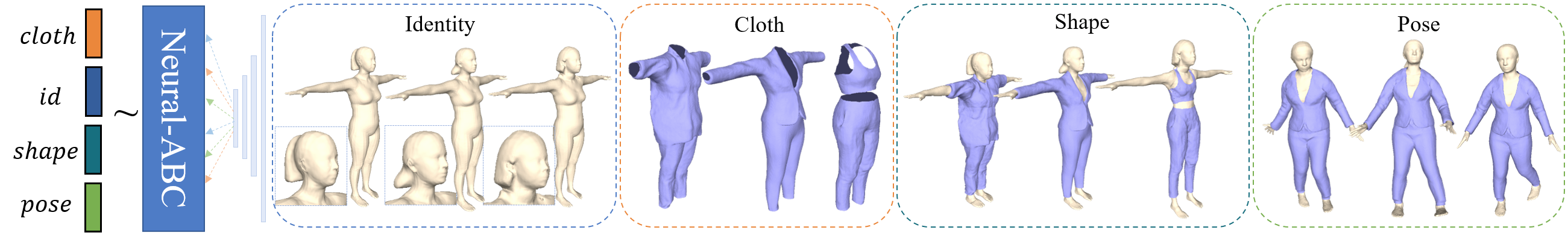}
  \caption{Neural-ABC is a neural implicit parametric model with latent spaces of human identity, clothing, shape and pose. It can generate various human identities and different clothes. The clothed human body can deform into different body shapes and poses.}
  \end{figure*}
  
\IEEEpeerreviewmaketitle

\IEEEraisesectionheading{\section{Introduction}\label{sec:introduction}}

\IEEEPARstart{P}{arametric} human body representation, which encodes human body attributes into low dimensional spaces, has wide applications in computer vision and computer graphics, such as human body reconstruction\cite{Kanazawa_2018_CVPR, plankers2001articulated, sminchisescu2006learning, li2022cliff}, pose estimation\cite{petrovich21actor, sminchisescu2002human}, personal avatar creation\cite{li2022avatarcap, zhi2020texmesh, saito2021scanimate, MetaAvatar:NeurIPS:2021, chen2022gdna}, and video editing\cite{zanfir2020human}.

By learning from a large amount of human body data, parametric human body representation is usually controlled by several latent variables, such as identity, pose, face, and hand, to represent various complex human body shapes. 
The constructed parameter model provides the possibility of decoupling the representation of geometric shapes through several compact low-dimensional latent codes.

Since the start of this century, parametric human body models have achieved remarkable success in representing specific human attributes, such as minimally clothed body and pose\cite{Anguelov05scape:shape, loper2015smpl, jiang2020disentangled}, hands\cite{romero2017embodied}, and separated clothing\cite{jiang2020bcnet, corona2021smplicit, ren2022dig, de2023drapenet}.
These models have been proven to have achieved significant results in representing the shape of minimally clothed bodies and have been widely applied in various applications. However, the parameterized representation of the clothed human body has always been an unresolved issue.

To unify the representation of the clothed human body, 
some methods based on SMPL+D\cite{pons2017clothcap,zhang2017detailed,alldieck2018video,ThiemoAlldieck2019Tex2ShapeDF,alldieck2019learning,bhatnagar2019mgn,ma2020learning,bhatnagar2020loopreg} add vertex displacement to the SMPL\cite{loper2015smpl} in the canonical pose, and then use linear blend skinning deformation to obtain the clothed human body geometry of the target pose.
This representation method is simple, but it is difficult to represent clothing that is inconsistent with the mesh topology of SMPL, such as skirts.
Implicit representation is very suitable for representing the human body due to its powerful representation ability.
\cite{deng2020nasa, saito2021scanimate, chen2021snarf, tiwari2021neural} construct pose representations for implicit geometry. 
\cite{mihajlovic2021leap, alldieck2021imghum} construct shape and pose representations for human body, excluding clothing.
NPMs\cite{palafox2021npms} and gDNA\cite{chen2022gdna} use a neural implicit function to represent the human body wearing clothes without being constrained by topology, but clothes and body cannot be separated by decoupled representation.

Another solution is to model the clothing and underlying body separately.
The body is generally represented by SMPL.
For clothing, separate body and clothing templates are used to register body motion sequences to obtain clothing geometry\cite{pons2017clothcap, yu2019simulcap}. Some methods\cite{jiang2020bcnet,corona2021smplicit,ren2022dig,de2023drapenet} train category independent representations based on the type of clothing.
BCNet\cite{jiang2020bcnet}, which models clothing as mesh, can obtain a unified representation of the clothed human body, and while independently controlling the clothing and underlying body, it is necessary to design topology for each type of clothing in advance.
\cite{corona2021smplicit, ren2022dig, de2023drapenet}use implicit representation for clothing modeling, bringing good topological flexibility to clothing representation. 
However, the body and clothing are not uniformly represented, which often requires special design when applied.
For example, SCARF\cite{Feng2022scarf} proposes special rendering to handle the human body represented by meshes and clothing represented by neural implicit functions. 
Therefore, a parameterized model that achieves the best of both worlds is needed: the flexibility of neural implicit functions to represent any clothing topology, as well as a unified geometric representation of the human body and clothing that can be independently controlled.

With this in mind, we propose Neural-ABC, a parameterized model for the clothed human body. 
It uniformly represents the clothed human body as deformable neural implicit functions, and parameterizes the clothed human body into disentangled identity, clothing, shape and pose.
We use an auto-decoder-architected neural implicit representation, which allows us to control the geometric shapes using low-dimensional latent spaces.
Specifically, to represent the complex topology of the clothed human body, we construct a cascading structure to establish a unified deformable unsigned distance function representation of the underlying body and clothing that can be disentangled.
For identity representation, the minimally clothed body is first represented as a signed distance field to ensure closure of the body.
Although the parameterized model of the minimally clothed body represented by the mesh has achieved significant results, the representation based on implicit functions is more advantageous for humans with head differences, as it does not require topological consistency with training data.
A reasonable solution is to model the clothing as an unsigned distance field\cite{ren2022dig,de2023drapenet,corona2021smplicit,guillard2022udf,zhao2021learning,long2022neuraludf,chibane2020ndf}.
To ensure a unified representation of clothing and body, we convert the body representation into an unsigned distance field.
We design the deformation of bodies and clothes to rely on the shape of body without affecting the geometric shapes.
Our method decouples the geometric shapes of the body and clothing, accommodating various body types.
Identity and clothing representation, modeling the canonical pose of the clothed human body. 
We use neural network-based pose representation and use linear blend skinning(LBS)\cite{loper2015smpl} to deform it to the specified pose.

We construct a training dataset, including a publicly available real scan dataset, two commercially available datasets, a publicly synthesized dataset containing different clothing,
and a new dataset synthesized through professional simulation software, which includes thousands of human shapes wearing different clothes and postures.
After constructing the parameter model, we can optimize the identity, clothing, shape and pose latent codes to fit new observations.
Due to the powerful and compact representative ability of the clothed human body, our model can achieve advanced results in fitting partial inputs, compared with other parametric human body representations. 
We show the representation ability of Neural-ABC by fitting the 3D human scans, depth maps and recovery from images of the human bodies wearing different clothes, and editing human attributes after fitting. 
Compared with the state-of-the-art parametric human body model, Neural-ABC has better representation ability in terms of clothing and semantic control, which provides a new method for the reconstruction and editing of the clothed human body. 
In summary, the contributions of this paper include the following:
\begin{itemize}
  \item We propose a parametric model based on the neural implicit function for the human bodies with clothes. 
  It achieves a unified and effective representation to support different clothing types without constraints by predefined templates.
  \item Compared with existing neural implicit parameter models, Neural-ABC represents the underlying human body and clothing uniformly, providing convenience for downstream tasks.
  \item We apply Neural-ABC to fit raw scans, depth maps and images, and the fitting results can be freely edited for identity, shape, clothing and pose.
  \item We construct a dataset containing decoupled information about the human body and clothing with clothing details. The dataset will be publicly available.
\end{itemize}

\section{Related Work}

\textbf{Parametric Human Body Model}.
Parametric human body models\cite{Anguelov05scape:shape,loper2015smpl,xu2020ghum, pavlakos2019expressive} represent the human body shape via groups of low dimensional parameterized representations by learning its statistical distribution from a large amount of human body data. 
SCAPE \cite{Anguelov05scape:shape} disentangles the human body shape into parametric representations of identity and pose for the first time. 
After that, SMPL\cite{loper2015smpl} and its variants parameterized the minimally clothed body\cite{hesse2019learning, osman2020star}, face\cite{pavlakos2019expressive,joo2018total}, hand\cite{romero2017embodied}, pose\cite{mahmood2019amass} and soft tissue\cite{santesteban2020softsmpl}, which became a very popular method. 
\cite{alldieck2021imghum} constructs the parameterized minimally clothed body model based on implicit functions.

Early parametric human body models mainly focus on constructing parametric representations of minimally clothed body bodies.
Meanwhile, many applications require the recovery of the geometric shape of the human body wearing clothes. 
However, the rich shapes and topologies of clothing make the geometric representation of the clothed human body much more difficult. 

\textbf{Clothing Model}.
A widely used dressing human body model is SMPL+D\cite{pons2017clothcap,zhang2017detailed,alldieck2018video,ThiemoAlldieck2019Tex2ShapeDF,alldieck2019learning,bhatnagar2019mgn,ma2020learning,bhatnagar2020loopreg}, which is designed to add vertex displacement under the SMPL canonical pose, and is deformed to obtain the clothed human body geometry of the target pose. 
This kind of representation is simple and compact, but it has limited representation ability due to the fixed topology of the underlying SMPL model. 
Therefore, it cannot represent loose clothes such as dresses. 

Another strategy is to model the clothing and the human body separately. The underlying body typically uses existing parameterized models, such as SMPL\cite{loper2015smpl}.
BCNet\cite{jiang2020bcnet} constructs mesh representations for different clothing types outside of SMPL.
DeepCloth\cite{deepcloth_su2022} uses UV maps with masks to represent the geometric shape of clothing.
\cite{corona2021smplicit, ren2022dig, de2023drapenet} use implicit unsigned distance fields to represent clothing, which has better flexibility compared to explicit representation.
Constructed clothing is inferred with the details of the human body’s motion information\cite{saito2021scanimate,li2022avatarcap}, or simulated by some physical simulation methods\cite{santesteban2021garmentcollisions,yang2018physics,su2020mulaycap}. 

Some methods use unsigned distance fields to represent clothing with open surfaces. 
\cite{corona2021smplicit, ren2022dig} constructs the parameterized model of clothing using UDF representation.
However, because the marching cube\cite{lorensen1987marching} cannot directly extract the zero level set of UDF, an offset needs to be added when extracting the garment mesh, and the extracted mesh also has thickness.
MeshUDF\cite{guillard2022udf} improves the issue with the Marching Cube, and Drapenet\cite{de2023drapenet} proposes a new clothing representation based on it\cite{guillard2022udf}.
These methods only construct parameterized models for clothing and do not include the underlying human body.

\textbf{Neural Implicit Representation for Clothed Human Body}.
Neural implicit functions have recently been widely used to represent geometric surfaces.
By regarding the neural network as an implicit function.
Neural implicit functions can theoretically encode complex surfaces without the need for predictive topological structures.

For a complete representation of the human body (wearing clothing or minimal clothing) reconstruction, SDF or occupancy is usually used.
PIFu and related works\cite{saito2019pifu, he2020geopifu, saito2020pifuhd, hong2021stereopifu, zheng2021deepmulticap, ijcai2022p218, xiu2022icon,huang2020arch} utilize neural implicit functions to reconstruct the detailed clothed human body from monocular or multiview images. 

Neural implicit functions can use low-dimensional vectors to represent high-dimensional geometric shapes, and change low-dimensional latent spaces to represent different high-dimensional shapes, which is very suitable for representing human body shapes and poses.
\cite{deng2020nasa, chen2021snarf,  mihajlovic2021leap, chen2022fastsnarf, dong2022pina} learns to represent the human body pose where the human body shape is represented via an implicit function. The constructed model can handle complex human poses, but cannot change the shape of the human body. \cite{chen2022gdna,palafox2021npms} learns the neural implicit function based representation for the human pose and clothed body, but clothing and the underlying body cannot be decoupled and controlled.

In our method, clothing and the underlying bodies are modeled as implicit unsigned distance field representations. 
The four latent spaces can independently generate human identity (head differences), shape (height, short, fat, thin), clothing and pose, and new identities, clothing, shapes and poses can be decoupled and generated by modifying latent spaces.

\section{Method}

\begin{figure*}[ht]
\centering
\includegraphics[scale=0.38]{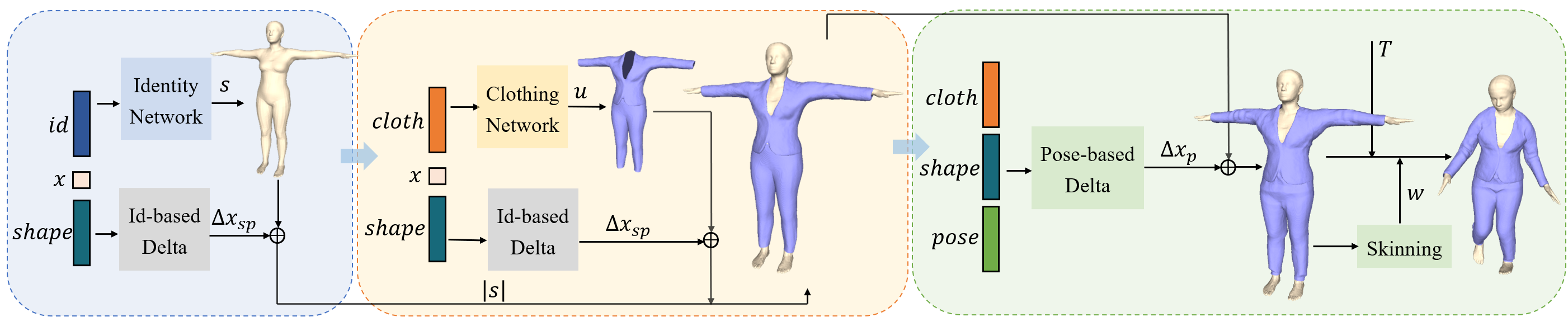}
\caption{Overview of our Neural-ABC pipeline. Given query points and latent codes, our model can generate the corresponding human body geometry by three modules.}
\label{fig_pp1}
\end{figure*}

In this section, we describe the proposed implicit parametric model (Neural-ABC) and the training strategies. 
Different from traditional human body representations, Neural-ABC achieves a unified representation of dressed human bodies wearing different types of clothing through neural implicit functions, 
and utilizes independently controllable identity and clothing parameterized representations to jointly represent the geometric shape of the clothed human body.
Based on an articulated structure, the dressed human bodies in the canonical pose can be transformed into various poses. 
To decouple the entanglement between identity and clothing, and improve the generalization ability of the trained model, we construct a cascaded representation of the human body and carefully design the training data and training strategy.
The four latent spaces can be jointly optimized to fit the new observations, and can be independently adjusted to modify the attributes of the human body. 
Figure~\ref{fig_pp1} shows an overview of our approach.

Section~\ref{overall} introduces the Neural-ABC model as a whole. In Section~\ref{id}, we discuss the identity and clothing representation, and shape represention in Section~\ref{shape}. Section~\ref{pose} introduces the pose representation. In Section ~\ref{loss}, we explain the loss functions employed in the training process.

\subsection{Neural-ABC Model}
\label{overall}
Our model utilizes four latent spaces to represent identity, clothing, shape and pose separately. 
We employ neural implicit representation for the identity space and encode the input points to the signed distance function (SDF) values to represent the basic shape of the minimally clothed body. 
The clothing space first uses unsigned distance fields to represent the average shape of the human body in clothing.
Clothing and minimally clothed bodies can be represented as unified unsigned distance fields.
The underlying body and clothing are then transformed to fit the specified body shape.

The pose space is defined as the axis angles of the joints similar to SMPL\cite{loper2015smpl}, and encodes the clothed body deformation flow from the canonical pose to the deformed pose. 
Therefore, for specified human identity and clothing, the existing action sequences\cite{mahmood2019amass} can be used to generate a large number of posed human bodies with clothes.

Given a dataset with different identities, clothing styles and poses, our goal is to learn a neural network-based implicit representation $f(\bm{x}, \bm{\alpha})$.
For a given set of latent codes $\bm{\alpha}=(\bm{\beta}_{i}, \bm{\beta}_{c}, \bm{\beta}_{s}, \bm{\theta})$, the geometry of any object \(\bm{M}\) can be obtained by $f$.
We construct four modules to represent identity $\bm{\beta}_{i}$, clothing $\bm{\beta}_{c}$, shape $\bm{\beta}_{s}$ and pose $\bm{\theta}$. 
Identity with shape and clothing with shape can be used together to represent the shape of the human bodies, and the pose space represents the articulated deformation from the canonical pose to the current pose.  
The zero iso-surface  $f(\bm{\cdot}, \bm{\alpha})=0$ can preserve all geometric details in \bm{$M$}, including identity, clothing, shape and pose.

\subsection{Identity and Clothing Representation}
\label{id}

The geometry of the dressed human body is determined by both the clothing and the underlying body.
We use the unsigned distance fields to represent the geometry of the dressed human body, as it can represent open clothing surfaces.
We define the identity space $\bm{\beta}_{i}$ to represent the human geometry (mainly including differences in the head) and the clothing space $\bm{\beta}_{c}$ to represent the clothes. 
The two spaces together determine the geometric shape of the clothed human body.

The minimally clothed body is first represented as a signed distance field, as the surface of the human body is closed.
The identity differences in the human body are primarily determined by the head.
Inspired by \cite{con_siren, Sitzmann_Martel_Bergman_Lindell_Wetzstein_2020}, we design the identity network as a conditional Siren.
The identity network predicts the implicit SDF for the different minimally clothed
bodies, to represent human identity. 
Each canonically-posed minimally clothed body $\bm{m}$ in the training dataset is encoded by ${D}_{i}$-dimensional latent shape code $\bm{\beta}_{i}$.
We implement our identity representation to predict the SDF values.
The query points $\bm{x}$ are encoded as signed distance values of the specified identity minimally clothed body, as, 
\begin{equation}
\begin{aligned}
f_{{\theta}_i}: {\mathbb{R}^3}\times{\mathbb{R}^{{D}_{i}}} &\mapsto{\mathbb{R}},
\\(\bm{x}, \bm{\beta}_{i}) &\mapsto f_{{\theta}_i} (\bm{x}, \bm{\beta}_{i}) = {\bm{s}}.
\end{aligned} 
\label{formula_id}
\end{equation}
This representation is similar to \cite{park2019deepsdf} but distinct from \cite{santesteban2021garmentcollisions}. Because the Equation (5) of \cite{santesteban2021garmentcollisions} generalize shape blend-shape offset values controlled by shape code, while our Equation~\ref{formula_id} predicts signed distance values.

For clothing representation,  the clothing geometric shape is determined by the clothing type itself (such as skirts or pants, and the length of the sleeves) and the shape of the underlying body, in the canonical pose.
The clothing type representation is defined in the mean shape space, consistent with the SMPL template body shape.
Each garment is encoded by the latent clothing code $\bm{\beta}_{c}$.
The network predicts the unsigned distance for a given point.
We train a unified clothing network for all types of clothing without the classification\cite{corona2021smplicit, ren2022dig, de2023drapenet}.
\begin{equation}
\begin{aligned}
  \label{equ_cloth}
f_{{\theta}_c}:{\mathbb{R}^3}\times{\mathbb{R}^{{D}_{c}}} &\mapsto{\mathbb{R}}, 
\\ (\bm{x}, \bm{\beta}_{c}) &\mapsto f_{{\theta}_c}(\bm{x}, \bm{\beta}_{c})=d
\end{aligned} 
\end{equation}

The signed distance value of the minimally clothed body and the unsigned distance value of the clothing can be unified.
For each given point of the mean shape, as well as the latent codes $\bm{\beta}_{i}$ and $\bm{\beta}_{c}$ of identity and clothing, the unsigned distance value of the clothed body can be uniformly calculated as,
\begin{equation}
\begin{aligned}
d(\bm{x}, \bm{\beta}_{i}, \bm{\beta}_{c})= \min(\left| f_{{\theta}_i} (\bm{x}, \bm{\beta}_{i})\right|, \left|f_{{\theta}_c}(\bm{x}, \bm{\beta}_{c})\right|)
\end{aligned} 
\end{equation}

Both the identity network and clothing network are trained in an autodecoder manner.
Each identity and clothing has an independent corresponding latent code, which is randomly initialized and jointly optimized during the training stage.

\subsection{Shape Representation}
\label{shape}
Deforming specified clothing and identity to a designated shape can be expressed as,
\begin{equation}
\begin{aligned}
  \bm{x}_{(\bm{\beta}_{i}, \bm{\beta}_{c}, \bm{\beta}_{s})} = \bm{x} + \bigtriangleup{x_s}(\bm{x}, \bm{\beta}_{s}),
\end{aligned} 
\end{equation}
$\bigtriangleup{x_s}(\bm{x}, \bm{\beta}_{s})$ is computed as in \cite{de2023drapenet, santesteban2021garmentcollisions}, which can only roughly fit the underlying body.
To make the deformation more accurate and adapt to more clothing in different shapes, we added synthetic data DressUp to train the deformation under partial clothing supervision, and more details are provided in~\ref{Dressup_data}.
During the training process, $\bm{x}$ are the generic 3D points which are independent of the identity or clothing type, so there is no need to retrain the shape representation when adding new clothing types or new identity.
During testing, for clothing, $\bm{x}$ are the template clothing vertices extracted by MeshUDF\cite{guillard2022udf} from Equation~\ref{equ_cloth}. For identity, $\bm{x}$ are the template identity vertices extracted by Marching Cubes\cite{lorensen1987marching} from Equation~\ref{formula_id}.

\subsection{Pose Representation}
\label{pose}
\begin{figure*}[ht]
  \centering
  \includegraphics[scale=0.45]{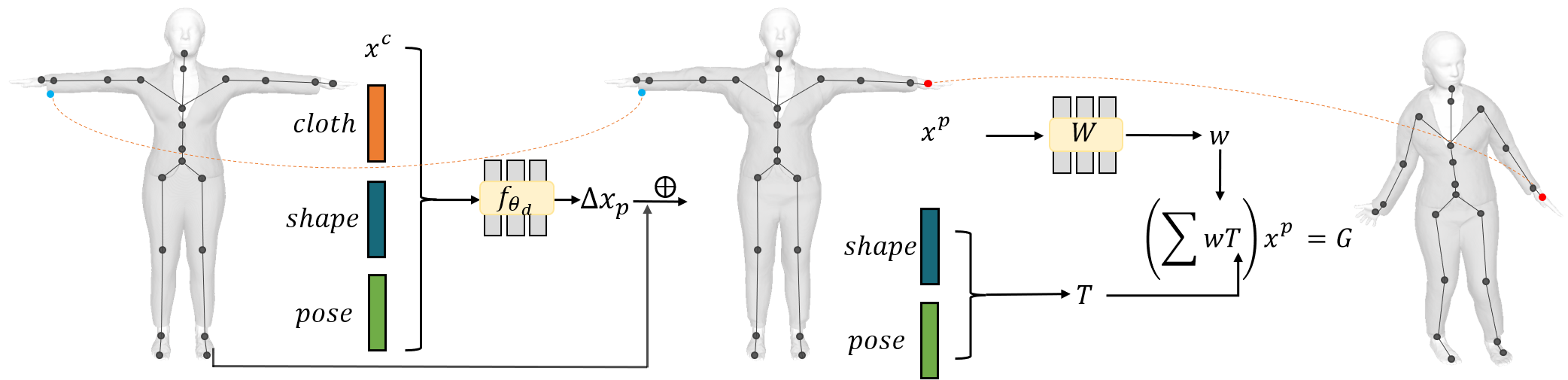}
  \caption{In the pose space, we use three latent spaces to encode query points in canonical pose. 
  First, calculate the pose dependent offset for all query points. 
  Then transfer them to the posed space by neural LBS.}
  \label{arch}
\end{figure*}

The pose space is used to represent the posed clothed bodies of different shapes. 
We decompose the pose deformation of the clothed human body into nonrigid deformation and linear blend skinning (LBS), as shown in Figure~\ref{arch}.

Given a point $\bm{x} \in {\mathbb{R}^{3}}$ in canonical space, the complete transformation of the dressed human body $G_b$ can be written as,
\begin{equation}
  \label{deform}
  \begin{aligned}
    G_b(\bm{\beta}_{i}, \bm{\beta}_{c}, \bm{\beta}_{s}, \bm{\theta}) &= W(\bm{x}_{(\bm{\beta}_{i},  \bm{\beta}_{c}, \bm{\beta}_{s}, \bm{\theta} )}, \bm{\beta}_{i}, \bm{\beta}_{c}, \bm{\theta}, \mathcal{W}(\bm{x})), \\
    \bm{x}_{(\bm{\beta}_{i},  \bm{\beta}_{c}, \bm{\beta}_{s}, \bm{\theta})} &= \bm{x} + \bigtriangleup{x_s}(\bm{x} + \bigtriangleup{x_p}(\bm{x}, \bm{\beta}_{i}, \bm{\beta}_{c}, \bm{\beta}_{s}, \bm{\theta}), \bm{\beta}_{s}) , \\
  \end{aligned} 
\end{equation}
where $W(\cdot)$ is the SMPL skinning funcion, applied with $\mathcal{W}(\bm{x})$.
$\mathcal{W}(\bm{x})$ is computed as in \cite{de2023drapenet, ren2022dig, santesteban2021garmentcollisions}.
The difference is that we deform the fully clothed human body, rather than just clothing.
Since $\mathcal{W}(\bm{x})$ does not require latent code as a condition, it can be directly used in unseen clothing without retraining.

We define the pose dependent nonrigid deformation in mean shape space, as $\bigtriangleup{x_p}(\bm{x}, \bm{\beta}_{i}, \bm{\beta}_{c},\bm{\beta}_{s}, \bm{\theta})$, predicted by $f_{{\theta}_d}$.
And then the clothing and identity are transformed into different body shapes by $\bigtriangleup{x_s}$.
We use linear blend skinning (LBS) to deform the points $\bm{x}_{(\bm{\beta}_{i},  \bm{\beta}_{c}, \bm{\beta}_{s}, \bm{\theta})}$ in the canonical pose space to the posed space.
The skinning weight is calculated as ${\bm{w}}=\mathcal{W}(\bm{x}_{(\bm{\beta}_{i},  \bm{\beta}_{c})})$.
Based on the kinematic tree, our model can maintain the articulated structure of the human body and independently control each joint. 
Training loss details can be found in Section~\ref{loss}.

\subsection{Loss Function}
\label{loss}

\textbf{Identity and Clothing Training.}
We train both the identity and clothing networks using loss functions as
  $L_\textrm{can}={\lambda}_{\textrm{d}} L_{\textrm{d}} + 
  {\lambda}_{\textrm{grad}}L_{\textrm{grad}} +
  {\lambda}_{\textrm{reg}}L_{\textrm{reg}} + {\lambda}_{\textrm{ek}}L_{\textrm{ek}}
  $.
$M$ is the number of identity objects to be trained in the canonical pose in the training dataset, and $N$ is the number of clothing objects.

For identity network, $I$ is the number of query points in each identity object, where $i=\{1,\cdots,I\}$ and $m=\{1,\cdots,M\}$.
$s$ is the signed distance value predicted by the identity network, and should be consistent with the ground truth $\overline{s}$, as, 
\begin{equation}
  \begin{aligned}
  L_{\textrm{d}}^{id}=\sum_{m=1}^{M} \sum_{i=1}^{I} \left \| s_{m,i}-\overline{s}_{m,i}\right \|_{1}. 
  \end{aligned} 
\end{equation} 

For clothing network, 
$J$ is the number of query points in each clothing object, where $j=\{1,\cdots,J\}$ and $n=\{1,\cdots,N\}$.
$d$ is the unsigned distance value predicted by the clothing network, and should be consistent with the ground truth $\overline{d}$, as, 
\begin{equation}
  \begin{aligned}
  L_{\textrm{d}}^{clo}=\sum_{n=1}^{N} \sum_{j=1}^{J} \left \| d_{n,j}-\overline{d}_{n,j}\right \|_{1}. 
  \end{aligned} 
\end{equation}

$L_{\textrm{grad}}$ is the gradient loss\cite{de2023drapenet}. 
For the clothing network, only the gradient of points with unsigned distance values other than 0 is calculated.
We use an L2-regularizer $L_{\textrm{reg}}$ on the zero-mean identity offset space $\bm{\beta}_{i}$ and $\bm{\beta}_{c}$, initializing as the Gaussian prior distribution, to enforce the compact shape manifold, as was found in \cite{palafox2021npms, park2019deepsdf}.
$L_{\textrm{ek}}$ is the eikonal loss\cite{icml2020_2086}.

\textbf{Pose Training.}
For pose, $O$ is the number of pose objects to be trained.
$K$ is the number of query points in the pose object, where $k=\{1,\cdots,K\}$ and $o=\{1,\cdots,O\}$.
We train $\bigtriangleup{x_p}(\bm{x}, \bm{\beta}_{i}, \bm{\beta}_{c}, \bm{\beta}_{s},\bm{\theta})$ by minimizing the loss as,
\begin{equation}
  \begin{aligned}
  L_{\textrm{p}} = {\lambda}_{\textrm{im}} L_{\textrm{im}} + {\lambda}_{\textrm{udf}} L_{\textrm{udf}} + {\lambda}_{\textrm{dis}} L_{\textrm{dis}} + {\lambda}_{\textrm{colli}} L_{\textrm{colli}}
  \end{aligned} 
\end{equation} 

$\bm{x}_{(\bm{\beta}_{i},  \bm{\beta}_{c}, \bm{\beta}_{s})}$ and $\bm{x}_{(\bm{\beta}_{i},  \bm{\beta}_{c}, \bm{\beta}_{s}, \bm{\theta})}$(abbreviated as $\bm{x}^c$ and  $\bm{x}^p$) are the corresponding points.
Since action sequences of the same shape have the same topology, the $V$ is the number of surface vertices in the training set, and can be used as landmark points, for the objects of action sequences, $v=\{1,\cdots,V\}$, 
\begin{equation}
\begin{aligned}
L_{\textrm{im}}=\sum_{o=1}^{O} \sum_{v=1}^{V} \left \| (\bm{x}_{o,v}^{c}+\bigtriangleup{x_p})-\bm{x}_{o,v}^{p}\right \|_{2}.
\end{aligned} 
\end{equation}

For any query point $\bm{x}^c$ and corresponding deformed point $\bm{x}^p$ in the pose space, the UDF value should be consistent.
$U(\bm{\cdot})$ calculates the UDF values for $K$ query points in the given clothing objects, 
\begin{equation}
  \begin{aligned}
  L_{\textrm{udf}}=\sum_{o=1}^{O} \sum_{k=1}^{K} \left \| U(\bm{x}_{o,k}^{c})-U(\bm{x}_{o,k}^{p})\right \|_{1}.
  \end{aligned} 
\end{equation} 

$L_{\textrm{dis}}$ is the L1-regularizer for the displacement $\bigtriangleup{x_p}$ of the corresponding points, to prevent excessive offset.
$L_{\textrm{colli}}$ prevents clothing from intersecting with the underlying body.
The $N$ objects of clothed bodies with $V$ surface vertices, as,
\begin{equation}
  \begin{aligned}
    L_{\textrm{colli}}= \sum_{n=1}^{N} \sum_{v=1}^{V} max(0, \epsilon-f_{{\theta}_i} (\bm{x}^{c}_{n,v}, \bm{\beta}_{i,n}))
  \end{aligned} 
\end{equation} 

Different from \cite{ren2022dig}, our constructed minimized clothed body parameterized model based on SDF can be directly used for collision detection without additional steps to convert the mesh into SDF. 
$\epsilon$ is a small value to prevent the points $\bm{x}$ in garment surfaces from overlapping with the body surfaces.
Therefore, collision loss can be directly used for representation training.

\section{Training Data Construction}
In this section, we describe how we collect data suitable for a pipeline, as well as the corresponding data preprocessing steps.
Our training set consists of two reprocessed publicly available datasets, two commercial datasets and an additional synthetic dataset that we constructed.
We apply several preprocessing steps to make the data is suitable for our human model learning.

\subsection{Data Requirements}
Parametric models require representational capabilities and decoupling capabilities of parametric spaces.
Our goal is to build a parameterized model that represents identity, clothing, shape and pose information for the human body.
To construct a model that can represent different human bodies wearing different clothes in different poses, the training data should cover the representation range of the model, for example, normal adult body shape and common clothing types.
We use the public real human body scan data and two commercial datasets, however, the existing available real data cannot provide a comprehensive dataset.
Therefore, we added additional synthetic data to enrich the training set.

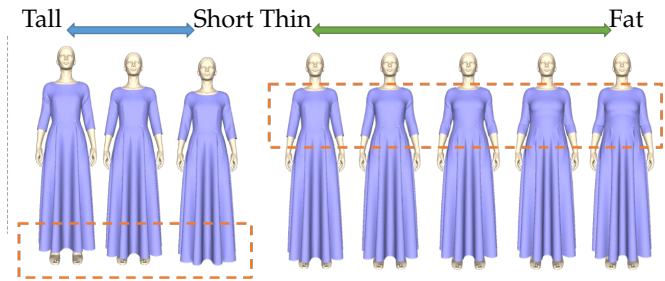
\begin{figure}[ht]
\centering
\includegraphics[scale=0.40]{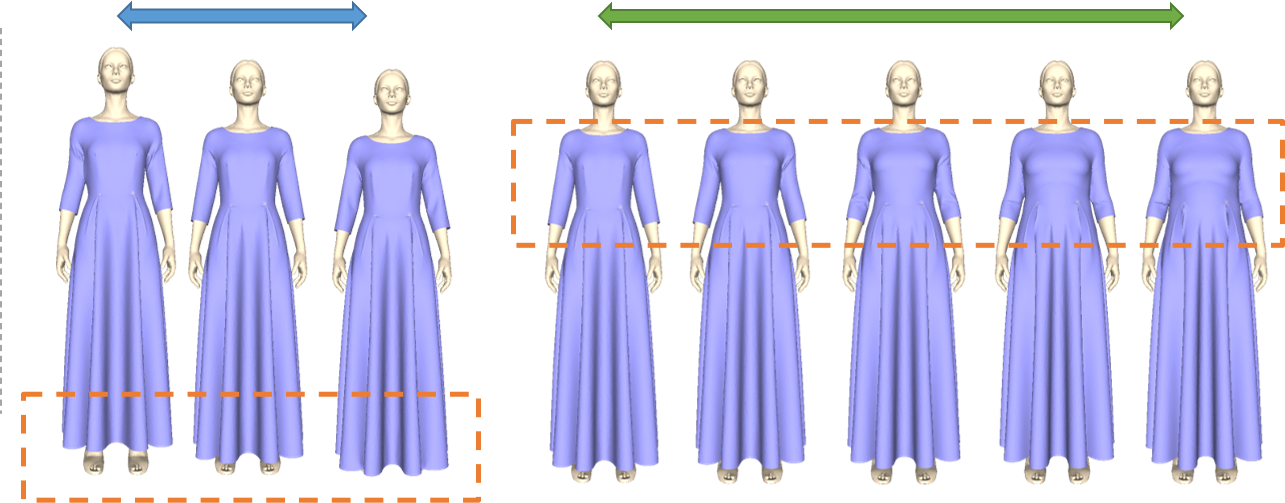}
\begin{picture}(0,0)
  \put(-245,95){Tall}
  \put(-180,95){Short}
  \put(-155,95){Thin}
  \put(-23,95){Fat}

\end{picture}
\caption{We show the objects of different body types wearing the same clothes in the DressUp dataset.
}
\label{thinfat}
\end{figure}

Data with decoupling information provide decoupling capability for the model.
For geometric shape in canonical pose decoupling, training data should include identity and clothing decoupling information.
It is not enough if each identity only wears unique clothes.
This is because different body types wearing the same clothes will have different effects.
Therefore, the dataset we built contains meshes with different identities wearing the same clothes in Figure~\ref{thinfat}, which provides identity and clothing decoupling information.
For pose decoupling, both shape in canonical pose and shape in arbitrary pose are required for each individual shape.
It is difficult to train the model using only the human mesh in an arbitrary pose. 
Many studies focus on obtaining its corresponding canonical pose from the mesh in an arbitrary pose\cite{chen2021snarf, chen2022fastsnarf}. 
This usually requires the a different pose sequences to obtain a reasonable canonical pose mesh.
Therefore, each individual shape we use includes the canonical clothed human body and the clothed human body in an arbitrary pose.
For the data without a canonical clothed human body, we use SMPL+D to register and transform the SMPL+D mesh to the canonical pose as the training agent of the clothed body in the canonical pose.

\subsection{Real Dataset}
\begin{figure*}[ht]
\centering
\includegraphics[scale=0.45]{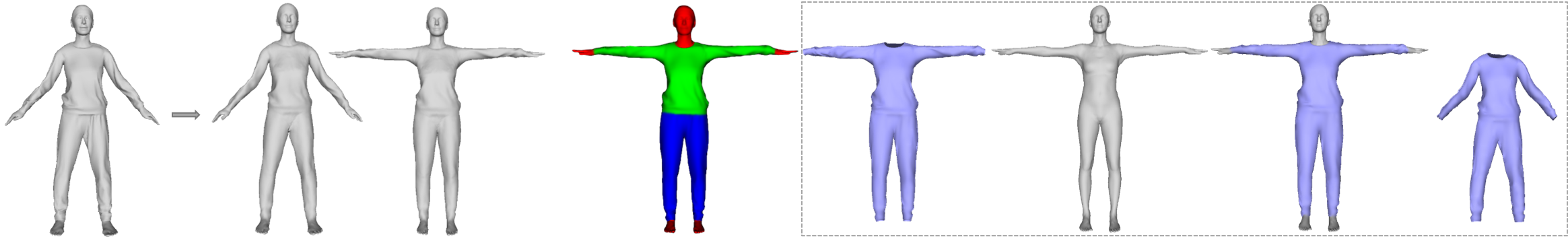}
\begin{picture}(0,0)
  \put(-485,-9){(a)}
  \put(-420,-9){(b)}
  \put(-372,-9){(c)}
  \put(-293,-9){(d)}
  \put(-225,-9){(e)}
  \put(-158,-9){(f)}
  \put(-88,-9){(g)}
  \put(-30,-9){(h)}
\end{picture}
\vspace{1mm}
\caption{The real scan (a) is first registered as SMPL+D (b), and then the SMPL+D in the canonical pose (c) is obtained. 
Using manually annotated segmentation information (d) can obtain separate clothing (e).
The SMPL in canonical pose (f) as a body agent for the clothed human body (g) for training.
(h) is the clothing in posed pace. 
The geometry inside the rectangular box is used to train our model.
}
\label{mgn_pro}
\end{figure*}
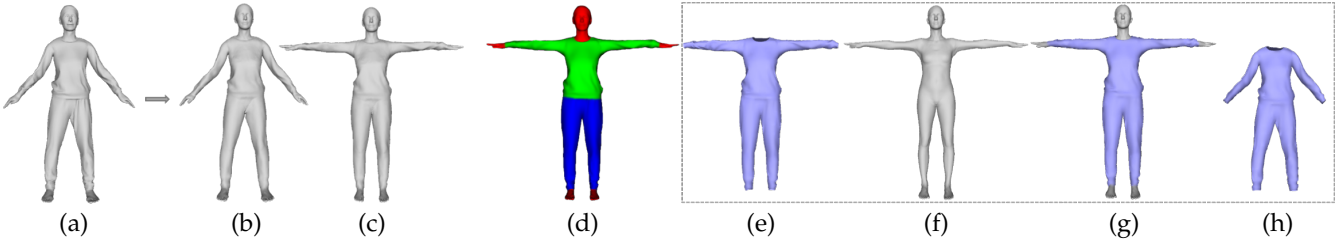

\begin{figure}[ht]
\centering
\includegraphics[scale=0.7]{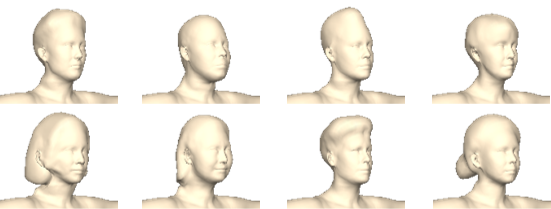}
\caption{Examples of underlying bodies with head variations.}
\label{head_data}
\end{figure}

We use real datasets, including publicly available datasets MGN\cite{bhatnagar2019mgn} and commercial datasets RenderPeople\cite{renderpeople} and AXYZ\cite{axyz}.
The MGN\cite{bhatnagar2019mgn} includes 94 independent subjects.
RenderPeople includes 97 independent subjects and AXYZ includes 181 independent subjects.
For each subject, we need to obtain a clothed human body in canonical space.
For the subject that provides skeleton and rigging, we directly transform it into canonical space.
For the subject that does not provide rigging, we use SMPL+D as a proxy to obtain the corresponding canonical human body mesh, by reverse LBS deformation.
Clothing is obtained through manually labeled segmentation information.
Since SMPL can only represent body shape, we cannot obtain accurate minimally clothed body details, 
so we only use clothing in canonical space and clothing in posed space for training,
and the SMPL in the canonical space is used as a proxy for the body to prevent collisions.
We show the processing steps in Figure~\ref{mgn_pro}. 
We select a hundred body geometries with head variations to represent identity differences in the underlying bodies.
We transfer the registered SMPL+D offsets in the head region to the mean shape. To ensure realistic deformation without truncation, we apply linear weights to the offsets in the neck region, with a weight of 1 at the upper end of the neck and 0 at the lower end. We showcase the head variations in the underlying bodies in Figure~\ref{head_data}.

\subsection{CLOTH3D Dataset}
\begin{figure}[ht]
\centering
\includegraphics[scale=0.55]{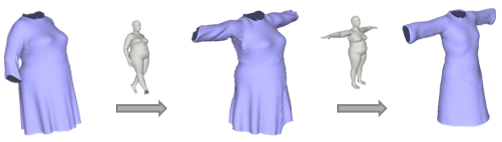}
\caption{
First, clothing with a pose utilizes the underlying SMPL to remove the pose, deforming it into a canonical pose garment. Then, the shape dependent deformation is eliminated, deforming it to the template shape.
}
\label{cloth3d_process}
\end{figure}

Due to the limited number of real datasets, we use the existing synthetic dataset to supplement more data, and reprocess it for training, to improve the generalization of the model.
We select objects from CLOTH3D\cite{bertiche2020cloth3d} and include 4048 independent subjects, with different SMPL shapes, clothes and pose sequences.
The dataset provides the clothed human body in canonical and other poses.
However, there is no clothed human body with the pose-blend-shape in the canonical pose.
Therefore, we use reverse LBS to transform clothes in different poses back to canonical pose space,
where skinning weight uses KNN to find the skinning weight on SMPL.
The meshes used for training: include minimally clothed bodies in canonical space, clothing in canonical space, clothed bodies in canonical space with deformation, and clothed bodies in posed space.
We show the processing steps in Figure~\ref{cloth3d_process}.

\subsection{Synthetic Dataset Construction}
\label{Dressup_data}
\begin{figure*}[ht]
\centering
\includegraphics[scale=0.195]{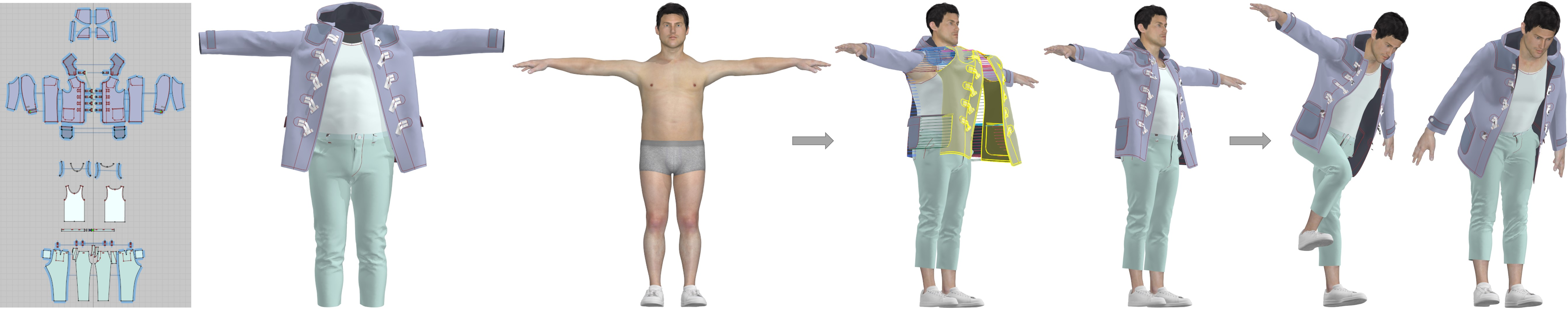}
\caption{Synthesis process of DressUp. Firstly, we prepare sewing patterns and virtual models. 
Then align the sewing pattern to the model.
Afterwards, wear clothes through physical simulation.
Finally, morph the model wearing clothes to the specified poses.}
\label{process}
\end{figure*}

\begin{figure*}[ht]
\centering
\includegraphics[scale=0.55]{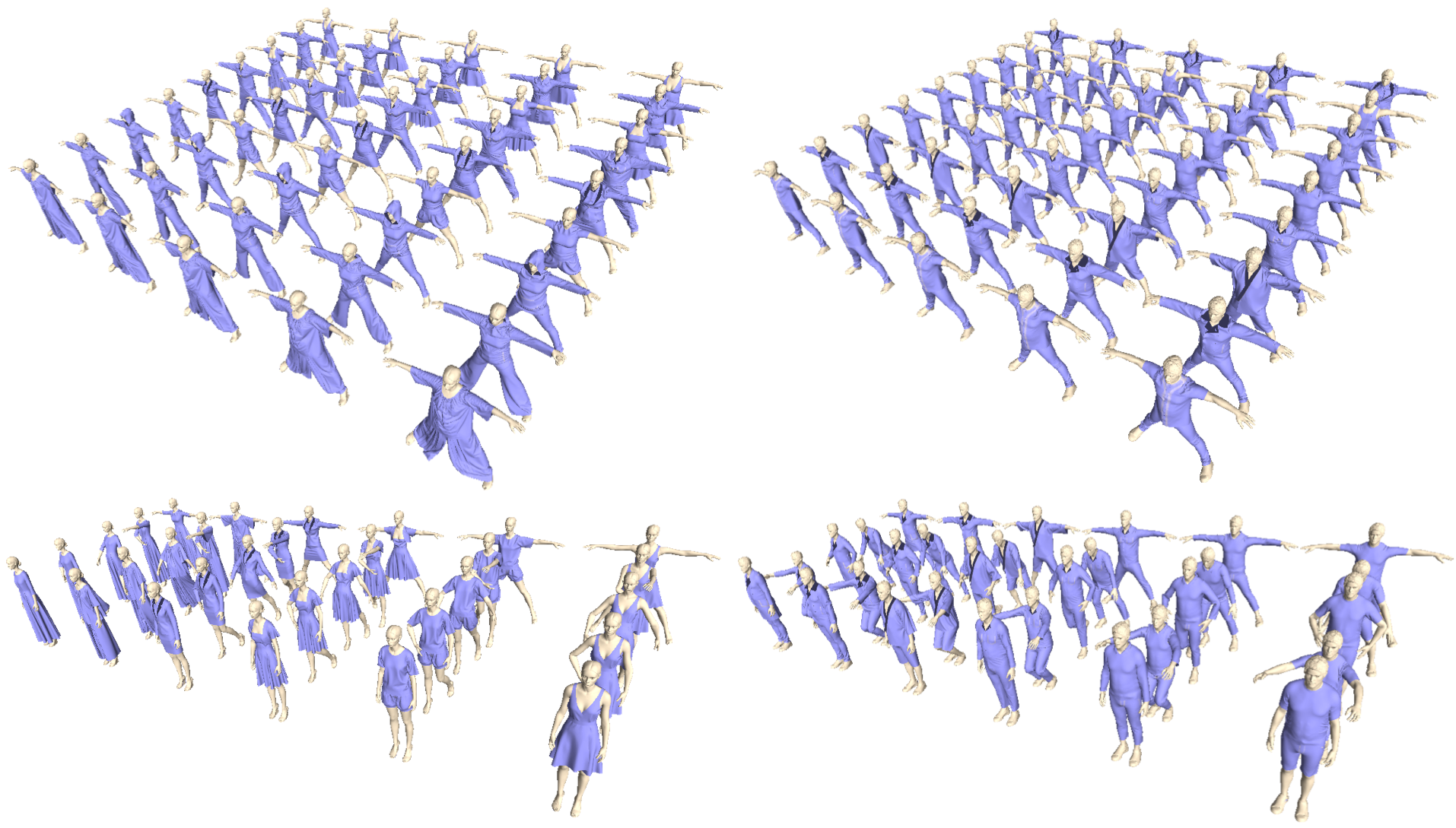}
\begin{picture}(0,0)
\put(-500,275){(a)}
\put(-500,120){(c)}
\put(-250,275){(b)}
\put(-250,120){(d)}
\end{picture}
\caption{The examples of our DressUp Dataset. (a)Women of different identities wear different clothes. 
(b) Men of different identities wear different clothes. (c) Women wear different clothes in different poses.
  (d) Men wear different clothes in different poses.}
\label{dataset2}
\end{figure*}
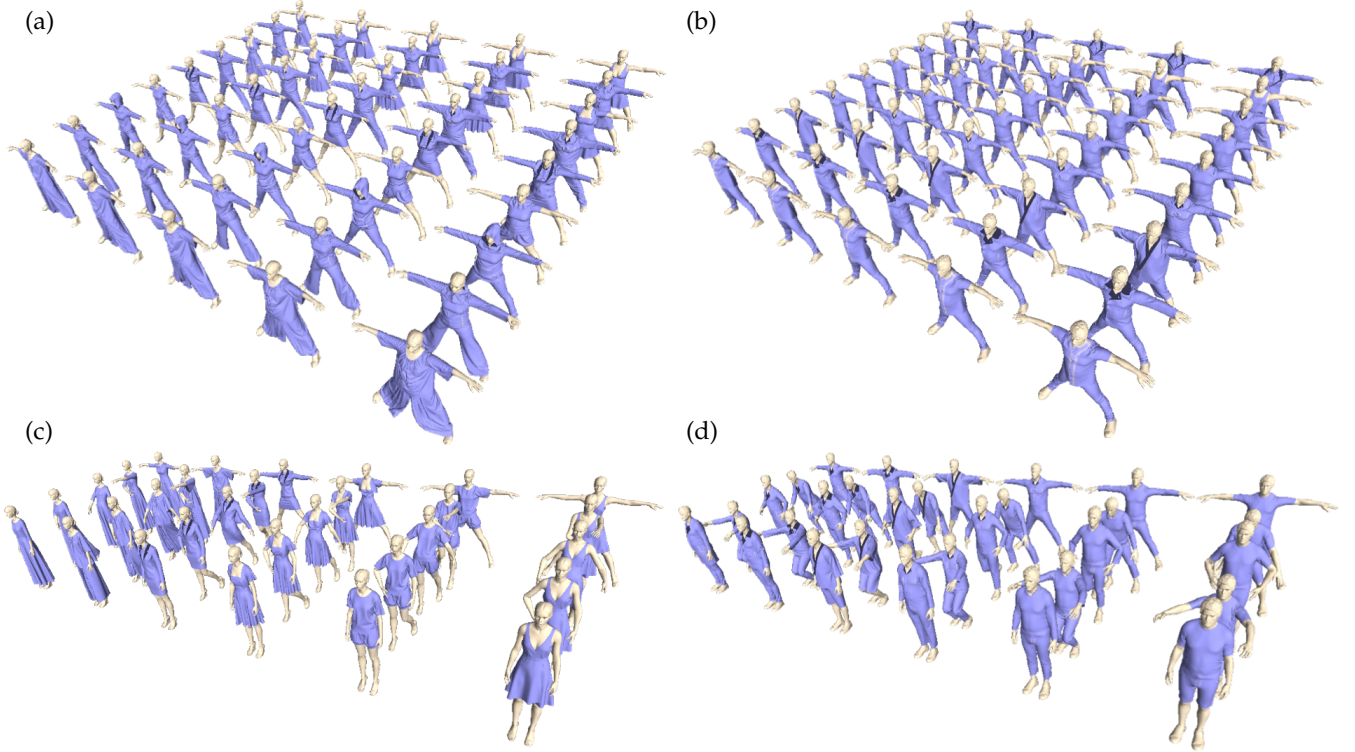

As there is currently no publicly available dataset that fully contains the decoupled information of the four latent spaces, and adds additional clothing types, we use physics-based simulation methods to construct our own dataset DressUp.
We use the professional clothing design software CLO3D\cite{clo3d} to construct the clothed body.
The dataset construction process can be divided into four steps: sewing pattern design, aligning clothing to the body, simulation, and retargeting the clothed body.
We show the data synthesis process in Figure~\ref{process}.
We design 30 male models and 30 female models with different body heights and widths. 
To maintain consistency, all male models and all female models have the same topology.
To increase the variety of clothing types, we manually design 67 sets of clothes, including T-shirts, trousers, shorts, coats, etc. 
Then we use the physical simulation to put clothes on the models.
To disentangle identity and clothing, we let models of different identities wear the same fitting clothes, and the same identity models wear different styles of clothes, then we obtain 1550 different shapes. 
Then the dressing models of T pose are animated to other poses, as shown in Figure~\ref{dataset2}. 
This dataset can be used to provide decoupling information for identity and cloth.

\subsection{Data Preprocessing}
\begin{figure*}[ht]
\centering
\includegraphics[scale=0.25]{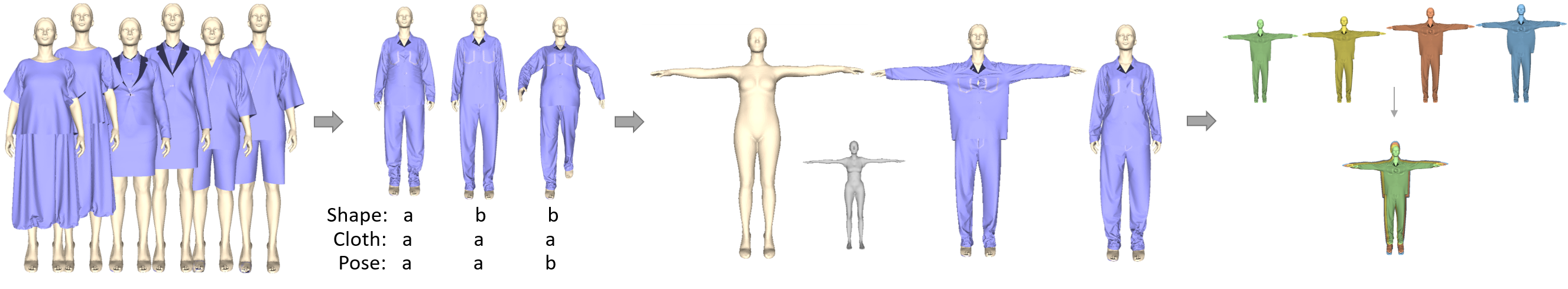}
\begin{picture}(0,0)
\put(-500,-12){Constructed dataset}
\put(-380,-12){Labeling}
\put(-250,-12){SMPL Registration}
\put(-90,-12){Data Alignment}
\end{picture}
\vspace{3mm}
\caption{The examples of data preprocessing. 
After constructing the original data set, we first label the different attributes of the each objects.
Then registering SMPL to minimally clothed body mesh.
Finally, all data are aligned according to the body.
}
\label{preprocess}
\end{figure*}

After preparing the raw data, it needs to be preprocessed to make it appropriate for training.
This includes attribute labeling, merging the clothing with the human body, registration of SMPL, and strict alignment of meshes to SMPL space.
In the following we elaborate on these steps. 
We show the data processing in Figure~\ref{preprocess}.

\textbf{Attribute Labeling.}
Since the attribute coupling in the data, for example, two objects have the same shape and different clothes.
We need to relabel the shared attributes.
For the DressUp dataset, the same shape is marked with the same shape number, and the same clothing is marked with the same clothing number, which corresponds to the latent codes during training.
For CLOTH3D and the real dataset, since there is no shared attribute, each number is independent.

\textbf{SMPL Registration.}
We register SMPL for the DressUp dataset.
For each object, we use its corresponding minimally clothed body data to register SMPL.
To make the registration results as accurate as possible, five points, including the nose tip, hands and feet, were manually marked as landmarks.
Because the body mesh has the same topology, it only needs to be marked once.

\textbf{Data Alignment.}
All data needs alignment.
Considering the independent representation  of clothing, we need to model the results of wearing the same clothes on different identities.
The position of clothes usually depends on the shoulders, and the position of pants usually depends on the crotch.
Therefore, we align the thigh root and arm root of all data.
We use registered SMPL to achieve alignment.
For each mesh $M$, we optimized the SMPL parameters, including global rotation $\bm{r}$ and translation $\bm{t}$.
The aligned new mesh is obtained by ${\bm{v}^{'} = \bm{r}^{-1}(\bm{v}-\bm{t})}$, where $\bm{v}$ are the points of $M$, and ${\bm{v}^{'}}$ are the points of aligned meshes.
We compare different alignment strategies in ~\ref{align}.

In addition, in both the CLOTH3D and real datasets, for each different clothing, only the geometry of the garment was worn on a different body shape.
Our clothing model is built in the standard body space of SMPL.
Therefore, we use SMPL as a proxy to transform clothing from its current body shape to the standard body shape.
For clothing of the current body shape, we calculate the nearest neighbor vertex from the clothing vertex $v_c$ to the SMPL vertex $v_b$.
Calculate the normal and offset from $v_c$ to $v_b$.
Then, clothing can be transferred at different SMPL corresponding vertices through normal and offset.
We transfer all clothing to the standard body shape and add Laplacian smoothing.

\section{Applications of Neural-ABC}
\label{sec:app}

Once training is completed, our representation can be regarded as a conditional decoder. 
The trained Neural-ABC can fit various types of inputs, by optimizing the identity, clothing and pose spaces.
After obtaining the corresponding latent codes, we can control the latent codes to adjust the clothed body shape.

\textbf{Fitting Neural-ABC to the depth maps.}
Here, we demonstrate how to fit Neural-ABC to depth maps.

Given the preprocessed identity code $\bm{\beta}_{s}^{0}$ and pose code $\bm{\theta}^{0}$,
we optimize $\bm{\beta}_{i}$, $\bm{\beta}_{c}$, $\bm{\beta}_{s}$, $\bm{\theta}$ at the same time, where $\bm{\beta}_{s}$ is initialized with $\bm{\beta}_{s}^{0}$, and $\bm{\theta}$ is initialized with $\bm{\theta}^{0}$.
We use DGCNN\cite{dgcnn,de2023drapenet} as the encoder for separate initialization of both $\bm{\beta}_{c}$ and $\bm{\beta}_{i}$.
In the canonical pose, for the given $\bm{\beta}_{i}$ and $\bm{\beta}_{c}$, MeshUDF\cite{guillard2022udf} can be used to extract meshes from implicit representations.
Because the process of extracting meshes from MeshUDF is differentiable, it can be optimized end-to-end.
The vertices of the extracted underlying body and clothing meshes are deformed using Equation~\ref{deform}.

We optimize the latent space by minimizing the following energy:
\begin{equation}
\begin{aligned}
L= {\lambda}_\textrm{cd}L_{\textrm{cd}} + {\lambda}_\textrm{reg}L_{\textrm{reg}} + {\lambda}_\textrm{ek}L_{\textrm{ek}}.
\end{aligned} 
\end{equation} 
$L_{\textrm{cd}}$ calculates the chamfer distance between the prediction mesh and ground truth.
$L_{\textrm{ek}}$ is the eikonal loss\cite{icml2020_2086}.
The following term is added to regularize the result:
\begin{equation}
\begin{aligned}
L_{\textrm{reg}}= {\lambda}_c \left \| \bm{\beta}_{c} \right \|_2 + {\lambda}_i \left \| \bm{\beta}_{i} \right \|_2 + {\lambda}_s \left \| \bm{\beta}_{s} - \bm{\beta}_{s}^{0} \right \|_2 + {\lambda}_p \left \| \bm{\theta}- \bm{\theta}^0 \right \|_2.
\end{aligned} 
\end{equation}
The model decoded by the final optimized codes $\bm{\beta}_{i}$, $\bm{\beta}_{c}$, $\bm{\beta}_{s}$, $\bm{\theta}$ is the fitting result.

\section{Experiments}

\subsection{Evaluations}

\begin{figure*}[ht]
\centering
\includegraphics[scale=0.24]{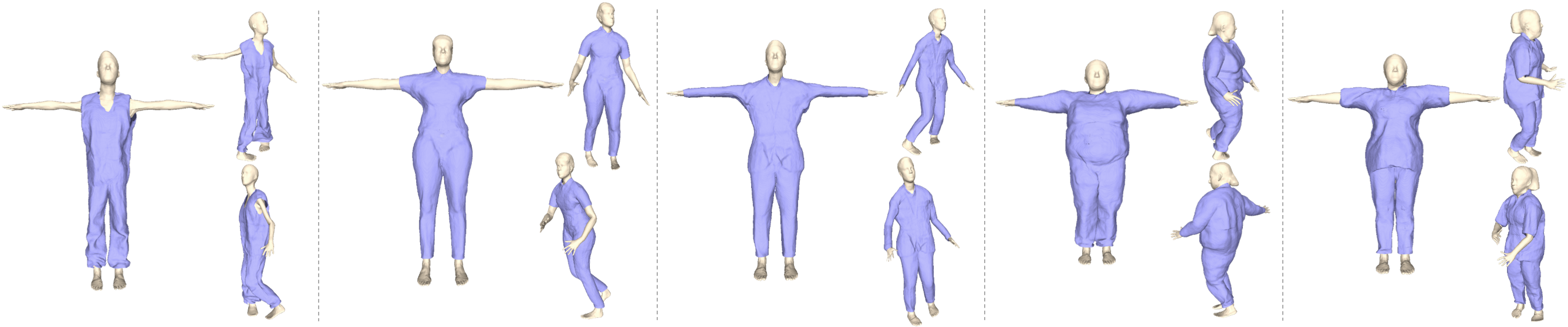}
\caption{We qualitatively show the generation results of Neural-ABC.
Left: clothed body in canonical pose.
right: clothed body in different poses.
}
\label{gen}
\end{figure*}

\textbf{Generation of Neural-ABC.}
We show the generation results of Neural-ABC in Figure~\ref{gen}. 
Our Neural-ABC can generate reasonable dressed bodies with posture.

\textbf{Disentangled Control.}
We show the decoupling ability of Neural-ABC to control the editing of corresponding properties by independently modifying the latent codes.
When fixing the clothing code, our model only fixes the style of the clothing, rather than fixing the geometry of the clothing as completely unchanged.
This is because clothing relies on the underlying body. 
For example, if the shape code is changed while fixing the clothing code, only the underlying body shape changes, which can cause the clothing to hang in the air or collide with the body.
Therefore, when only modifying the identity code, the clothing will follow reasonable changes in the underlying body without changing the clothing style.
Similarly, the geometry of the clothed human body in the current pose is determined not only by its poses, but also by the geometry of the clothed human body in the canonical space.
We showcase the results in the video.
The decoupling ability is reflected in modifying any parameter in the parameterized space, and the geometry represents only changes in that attribute, while other attributes remain unchanged.

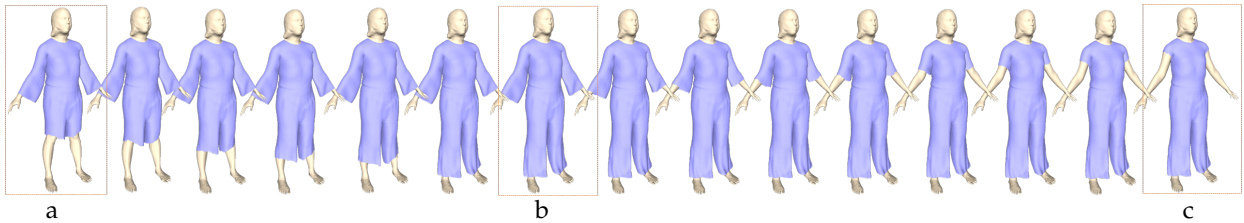
\begin{figure*}[ht]
\centering
\includegraphics[scale=0.38]{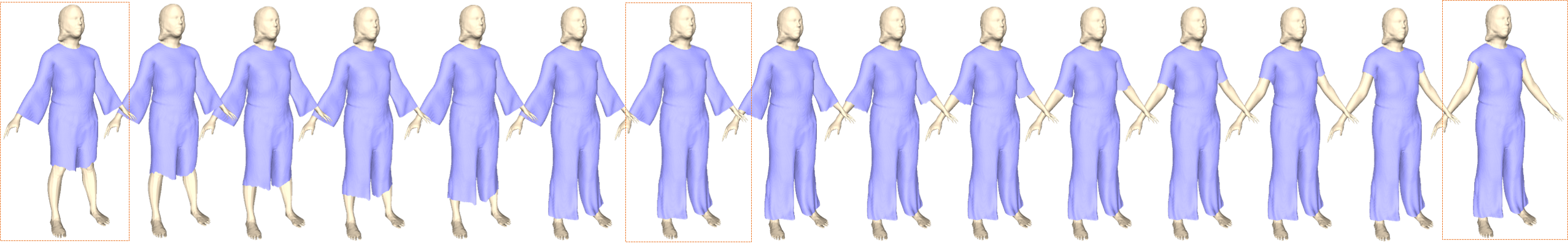}
\begin{picture}(0,0)
  \put(-455,-8){a}
  \put(-270,-8){b}
  \put(-25,-8){c}
\end{picture}
\caption{We show the continuity in the clothing space. From a to b, the length of the pants is interpolated.
  From b to c, the sleeve length is interpolated.
}
\label{interpolation}
\end{figure*}

\textbf{Latent Space Interpolation}. 
Our latent spaces of identity, clothing, shape and pose can be interpolated to obtain continuous generation results. 
This suggests the continuity of our model. We show the clothing space interpolation in Figure~\ref{interpolation}. 
By interpolating the clothing space, the style of clothing can be continuously changed. 
Although only the initial and final results are specified, the interpolation results can still maintain a reasonable shape.

\begin{figure}[ht]
  \centering
  \includegraphics[scale=0.26]{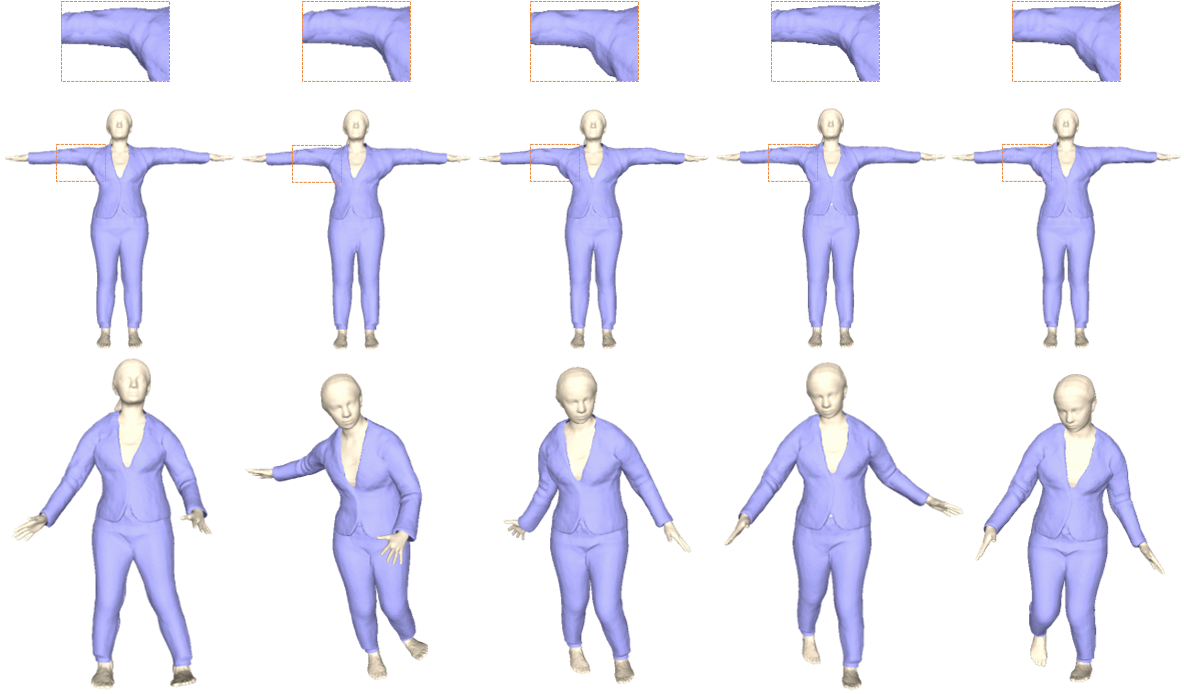}
  \begin{picture}(0,0)
    \put(-240,130){(a)}
    \put(-240,40){(b)}
  \end{picture}
  \caption{The animation results of the clothed human body.
    (a) We show the pose dependent deformation in canonical space.
    (b) Animation results driven by pose sequence.
  }
  \label{poseblend}
\end{figure}
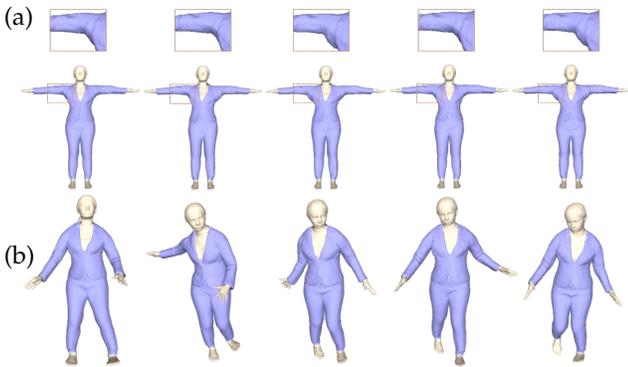

\textbf{Animation.}
\label{Animation}
  The generated clothed human body can use motion sequences to drive the skeleton. This is driven by modifying the pose code.
  We show the results in Figure~\ref{poseblend}. 
  In (a), the clothing worn exhibits subtle deformations.
  (b) Geometry of the dressed human body in different poses.

\begin{figure}[ht]
  \centering
  \includegraphics[scale=0.33]{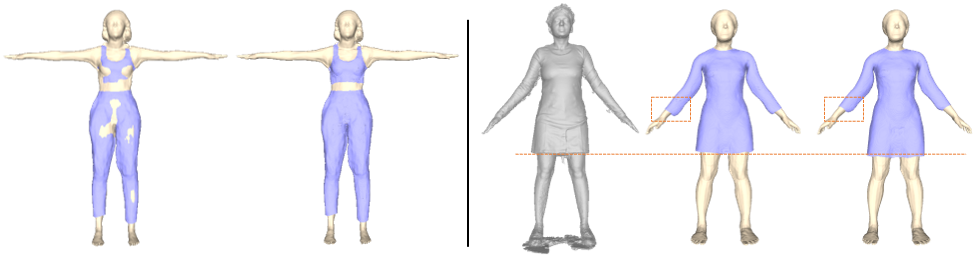}
  \begin{picture}(0,0)
    \put(-190,-7){(a)}
    \put(-70,-7){(b)}
  \end{picture}
  \caption{Ablation study on models trained with or without the DressUp dataset. 
  (a) The generation results of unseen clothing and body. Left: The model without the DressUp dataset. Right: The model with the DressUp dataset.
  (b) Left: Ground truth meshes. Middle: The model with the DressUp dataset. Right: The model without the DressUp dataset.}

  \label{ablation_new_data}
\end{figure}
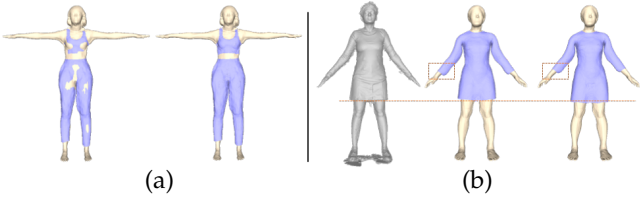

\textbf{Ablation Study on the DressUp Dataset.}
\label{Dressup_exp}
The DressUp dataset we constructed provides decoupling information between clothing and the underlying body.
It has improved the representation ability of the model.
We show the results in Figure~\ref{ablation_new_data}.
Our model needs to have the ability to wear designated clothing on different underlying bodies.
Clothing relies on the deformation effect of body shape, which comes from the cascade structure and the data-driven design.
We demonstrate the results of wearing designated clothing on an unseen body shape, in (a).
Without adding collision handling, the clothing deformation of the model without DressUp is not accurate, which leads to artifacts in the clothed body.
On the other hand, the DressUp dataset contains rich and more complex clothing compared to the CLOTH3D dataset, which can improve the representative power of our model.
We compare the models with or without the DressUp dataset by fitting the results.
Even if the deformation of clothing depending on body shape is inaccurate and can be alleviated through collision loss, the representation ability of the model still affects the fitting results.
In (b), the fitting results of the model with DressUp are more accurate in both skirt length and sleeve length.

\textbf{Ablation Study on Data Alignment.}
\label{align}
Data alignment affects the speed and accuracy of model training.
In canonical space, align the bodies (especially arms and legs) of the objects as much as possible to make it easier to wear clothes.
Therefore, we compared the strategies of aligning the feet\cite{hong2021stereopifu} and aligning the body.
Quantitative results are presented in Table~\ref{align}.

\begin{table}
\centering
\caption{Quantitative comparison of data alignment.  Numeric units: $(10^{-3})$}  

\begin{tabular}{cccc}
\toprule    

   & GT-to-Res $\downarrow$ & Res-to-GT $\downarrow$ \\
\midrule 
Identity + FA & 8.48 & 12.6 \\
Identity + BA & 6.32 & 9.10 \\
Clothing + FA & 8.23 & 11.4 \\
Clothing + BA & 6.14 & 8.52 \\
\bottomrule 
\end{tabular}
\label{align} 
\end{table}

\begin{figure}[ht]
  \centering
  \includegraphics[scale=0.32]{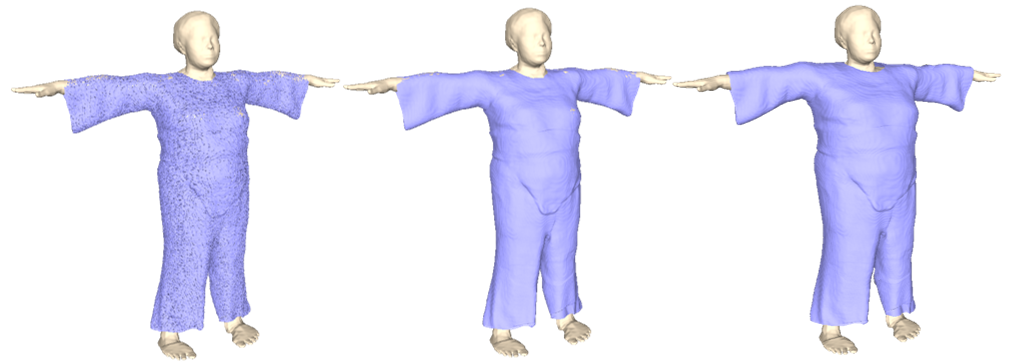}
  \begin{picture}(0,0)
    \put(-230,-10){w/o $L_{\textrm{colli}}$, $L_{\textrm{udf}}$}
    \put(-135,-10){w/o $L_{\textrm{colli}}$}
    \put(-50,-10){complete loss}    
  \end{picture}
  \caption{Comparison of model generation results without collision loss and landmark loss, without collision loss and complete loss during training.}
  \label{collision}
\end{figure}
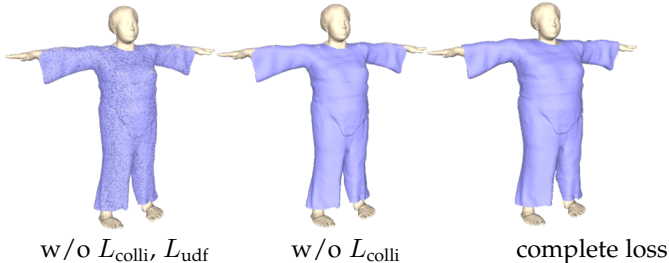

\textbf{Ablation Study on Pose Training Loss.}
Utilizing $L_{\textrm{udf}}$ can eliminate the roughness generated solely by $L_{\textrm{lm}}$, as $L_{\textrm{lm}}$ only employs surface vertices.
Using collision loss during training can eliminate collisions between clothing and body in the training set, as shown in Figure~\ref{collision}.
However, for the combination of unseen clothing and the body, there is still a possibility of collision.
Therefore, after extracting the mesh, for areas where clothing and body intersect, clothing can be directly deformed to the outside of the body, by moving clothing vertices a small distance along the normal away from the body.
This deformation is generally small, and the clothing will not be deformed to the wrong side of the body.

\subsection{Comparison}
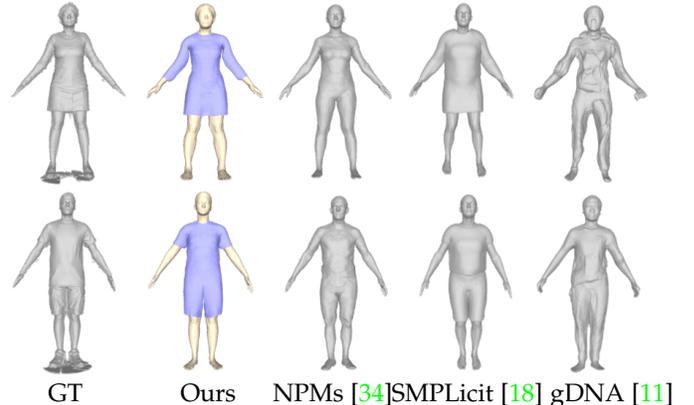
\begin{figure}[ht]
  \centering
  \includegraphics[scale=0.57]{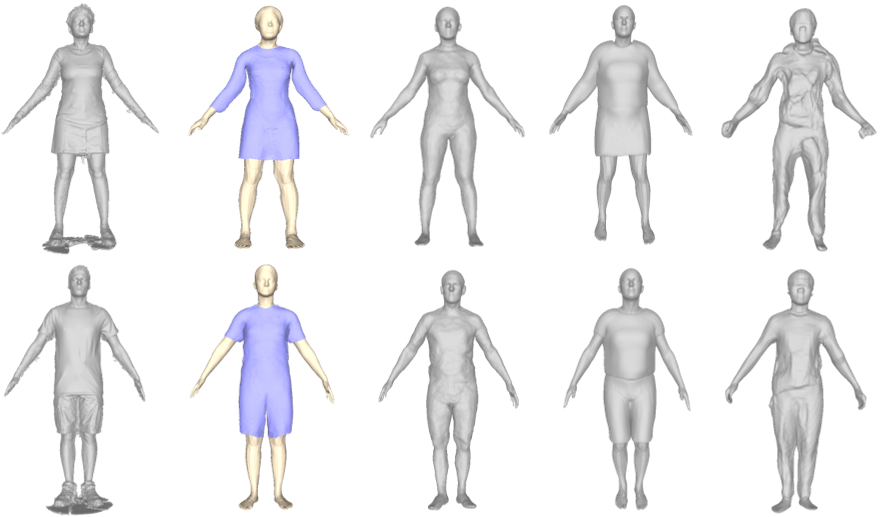}
  \begin{picture}(0,0)
    \put(-230,-10){GT}
    \put(-180,-10){Ours}
    \put(-145,-10){NPMs\cite{palafox2021npms}}
    \put(-100,-10){SMPLicit\cite{corona2021smplicit}}
    \put(-40,-10){gDNA\cite{chen2022gdna}}
  \end{picture}
  \caption{Comparison results with other state-of-the-art methods for fitting scans.
 }

  \label{Fit_scans}
\end{figure}

\textbf{Comparison with the Clothed Human Body Representation Model.}
Following\cite{corona2021smplicit}, we use the Sizer dataset\cite{tiwari20sizer} to evaluate the fitting.
All the compared models and our method are not trained on this dataset.
We perform a comprehensive comparison with state-of-the-art methods on different clothed human bodies, and show the clothing and pose editing capabilities on the new observations.
We evaluate the performance of the reconstruction results by chamfer distances.
We compare our model with the state-of-the-art deep-learning-based SDF implicit models SMPLicit\cite{corona2021smplicit}, NPMs\cite{palafox2021npms} and gDNA\cite{chen2022gdna}.
We provide additional clothing type segmentation information for SMPLicit.
The qualitative results are shown in Figure~\ref{Fit_scans}, and the quantitative results are shown in Table~\ref{tab_sacn}. 
Because the underlying bodies of the original scans are not visible, our results are waterproofed to calculate the error.

The fitting results of NPMs\cite{palafox2021npms} are close to minimally clothed bodies and lack clothing information. 
SMPLicit\cite{corona2021smplicit} can reconstruct the correct clothing type, but it needs additional clothing category and segmentation information. 
For gDNA\cite{chen2022gdna}, we compared it with the publicly available model. Our Neural-ABC can achieve the best fitting results on different types of clothing, and the underlying bodies include hair.

\begin{figure*}[ht]
  \centering
  \includegraphics[scale=0.47]{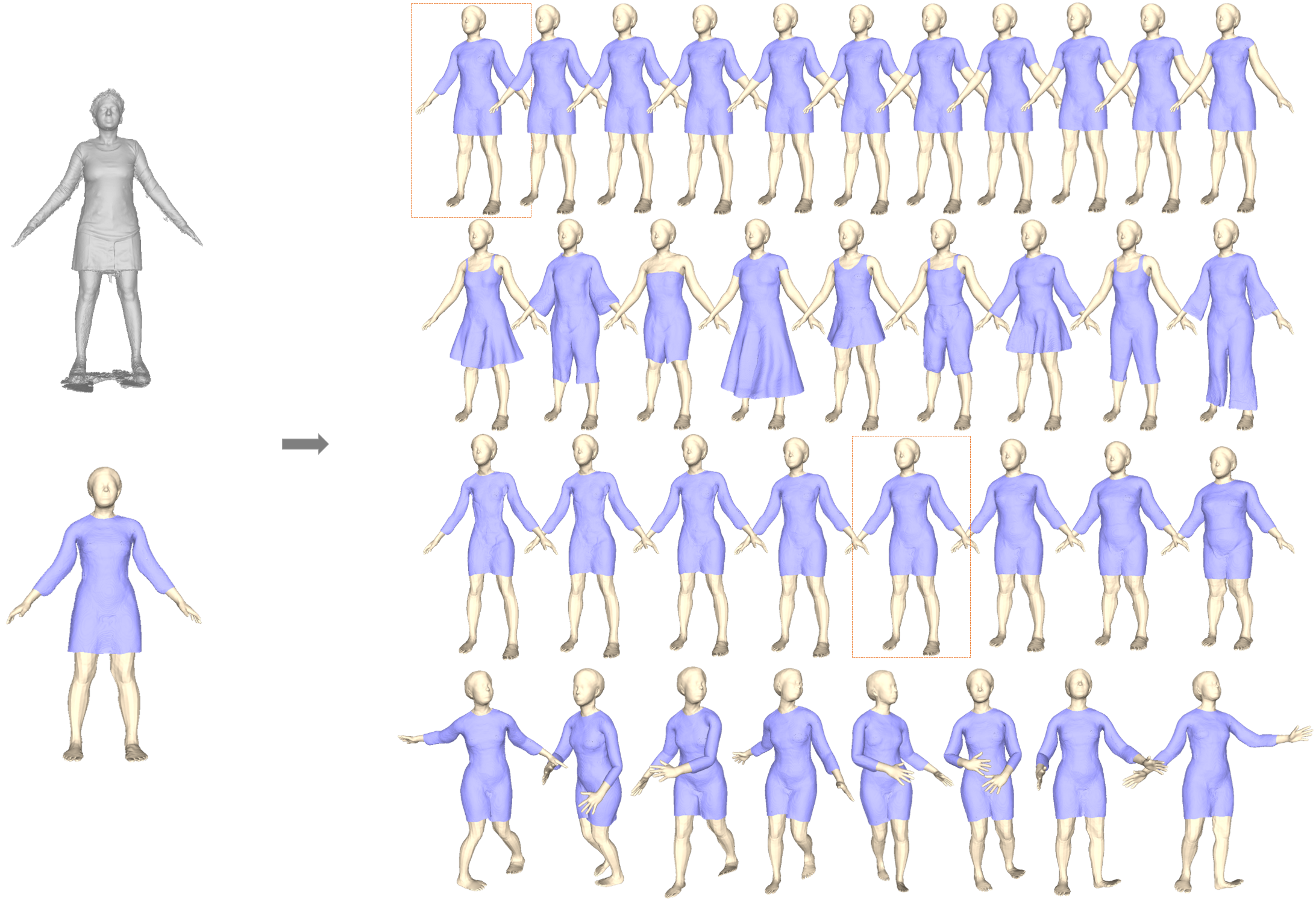}
  \begin{picture}(0,0)
    \put(-485,20){Fitting result}
    \put(-200,-9){Edit results}
    \put(-420,300){Change clothing 1}
    \put(-420,220){Change clothing 2}
    \put(-420,130){Change shape}
    \put(-420,40){Change pose}
  \end{picture}
  \caption{For the fitting result, the attributes can be edited by modifying different attribute codes. 
  We show the results of changing clothing, shape and pose codes.
  The rectangular boxes indicate the fitting results.}
  \label{Fit_change}
\end{figure*}
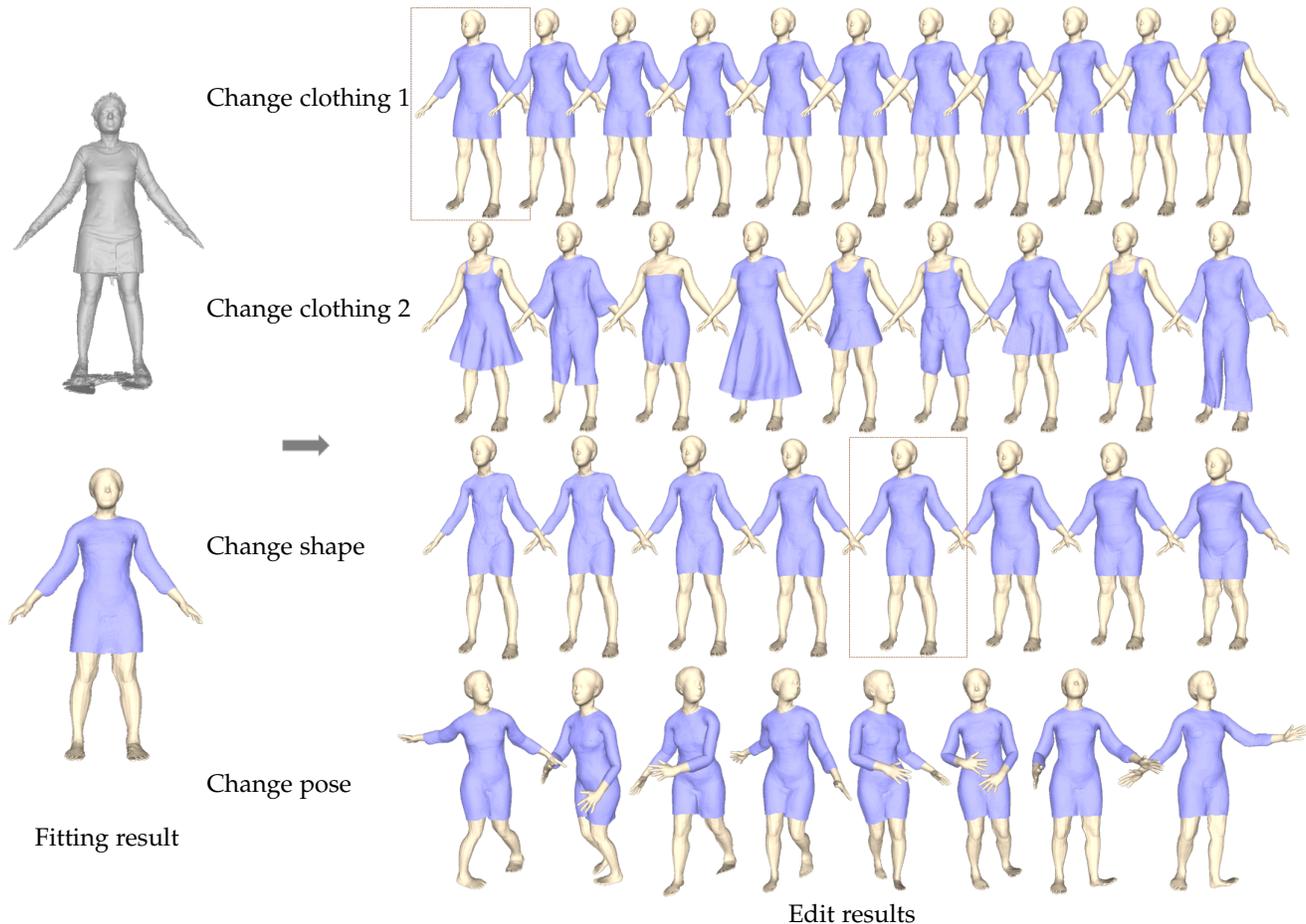

Since Neural-ABC decouples the parameterized spaces, the fitted results can edit attributes independently.
We show the results of editing clothing and pose for fitting results in Figure~\ref{Fit_change}. 
For changing clothing 1, the clothing code is changed continuously, allowing for continuous editing of sleeve length, while maintaining the same pose and underlying body shape.
For changing clothing 2, the clothing code is changed randomly, and the clothing is modified to a different style.
For changing shape, by only modifying the shape code, it is possible to modify the underlying body shape while maintaining the same clothing style and pose. 
Although the clothing code remains unchanged, the clothing results can still follow the changes in the underlying body and maintain correct wearing.
For changing pose, when specifying clothing and shape, we can modify the pose code to adjust the poses of the clothed human body.

\begin{table}
\centering
\caption{Fitting Comparison. We report the average distance (cm) between the reconstruction results and GT meshes.
}
\begin{tabular}{ccc}
\toprule   

Subject & GT-to-Res $\downarrow$ & Res-to-GT $\downarrow$ \\
\midrule 
NPMs & 2.12 & 1.63 \\
SMPLicit & 2.32 & 1.88 \\
gDNA & 1.92 & 1.44 \\
\textbf{Ours} & \textbf{1.12} & \textbf{1.21} \\ 

\bottomrule 
\end{tabular}

\label{tab_sacn} 
\end{table}

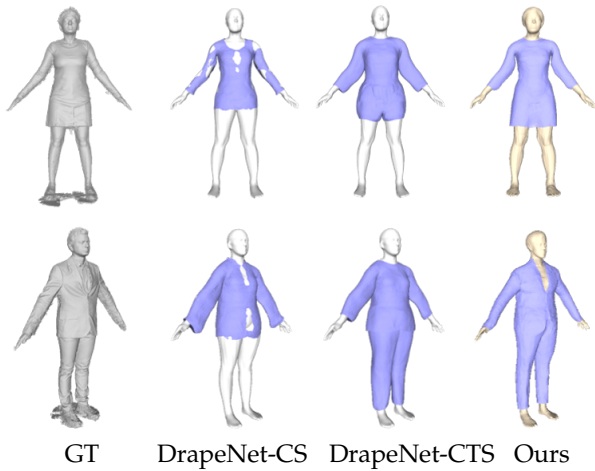
\begin{figure}[ht]
  \centering
  \includegraphics[scale=0.4]{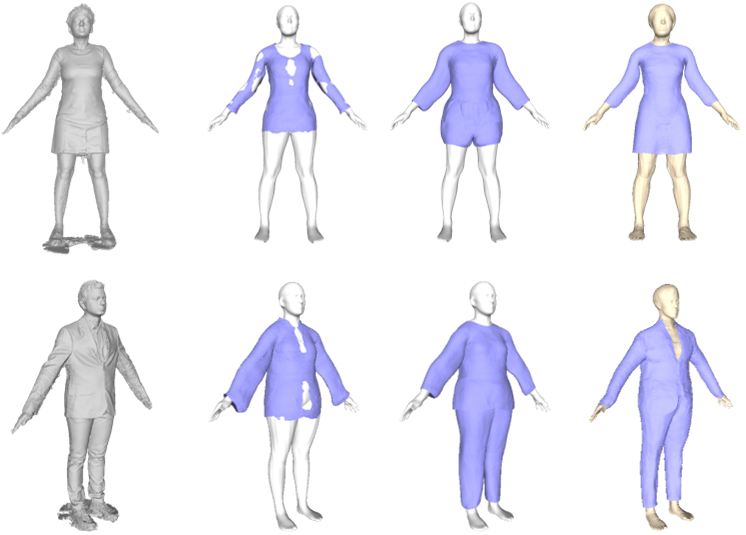}
  \begin{picture}(0,0)
    \put(-205,-12){GT}
    \put(-170,-12){DrapeNet-CS}
    \put(-105,-12){DrapeNet-CTS}
    \put(-35,-12){Ours}
  \end{picture}
  \vspace{2mm}
  \caption{Comparison results with DrapeNet\cite{de2023drapenet}.
   DrapeNet-CS: DrapeNet-Clothing Segmentation, using segmentation that only distinguishes between clothing and the body.
   DrapeNet-CTS: DrapeNet-Clothing Type Segmentation, using clothing segmentation that distinguishes between top and bottom clothing.
 }
  \label{compare_drapenet}
\end{figure}

We present the comparison results with the fitting Sizer dataset of DrapeNet\cite{de2023drapenet}, which is a UDF-based clothing model, and also use CLOTH3D\cite{bertiche2020cloth3d} as the training set, in Figure~\ref{compare_drapenet}.
During the fitting process, we optimized only the latent codes.
Because DrapeNet trains upper and lower clothing separately, when compared with DrapeNet, we only use segmentation to distinguish between clothing and the body(DrapeNet-CS), as well as further segmentation to distinguish between upper and lower clothing(DrapeNet-CTS).
DrapeNet-CS has difficulty representing complete human clothing.
DrapeNet-CTS represents the complete clothing, but it is not accurate.
Our method can accurately fit different types of clothing, using only segmentation that distinguishes between clothing and the body.
Additionally, DrapeNet uses SMPL as the underlying body, but it cannot represent information about the head.

\subsection{Applications}
\label{appli}
Compared with other parameterized representations of humans, our Neural-ABC models independent clothing representations, showing better performance in some human body reconstruction tasks.
In this part, we show the effect of Neural-ABC on fitting depth maps and images, and editing the clothing and pose of the fitting results.

\textbf{Fitting Neural-ABC to the Depth Map.}
We test our model in the public dataset ReSynth\cite{POP:ICCV:2021}, which provides human data wearing various types of garments in different poses.
All compared models and our method are not trained on this dataset.
We perform a comprehensive comparison with state-of-the-art methods on different clothed human bodies.
We reconstruct the ground truth meshes from point clouds in ReSynth by Poisson surface reconstruction\cite{kazhdan2006poisson}. 

We evaluate the performance of the reconstruction results by the intersection over union (IoU) and the chamfer distance from the ground truth meshes to the reconstruction result. 
The IoU measures the overlap between the predicted meshes and the ground truth meshes.
The distacne measures the accuracy of reconstructed surfaces.

Our method can reconstruct partial input into a complete human body, by optimizing only the latent codes.
We show the results of monocular depth map reconstruction, and compare them with those of  SMPLicit and NPMs.
We manually annotate the garment segmentation and clothing labels for SMPLicit, and optimize the SMPL parameters using the body point clouds for SMPLicit, instead of estimating SMPL parameters from the depth map. 
When we optimize, each posed point is projected and falls into a semantic segmentation pixel matching its garment types.
We adopt the same optimization reconstruction loss in Neural-ABC to make the reconstruction surfaces close to the depth maps.

\begin{figure*}[ht]
  \centering
  \includegraphics[scale=0.7]{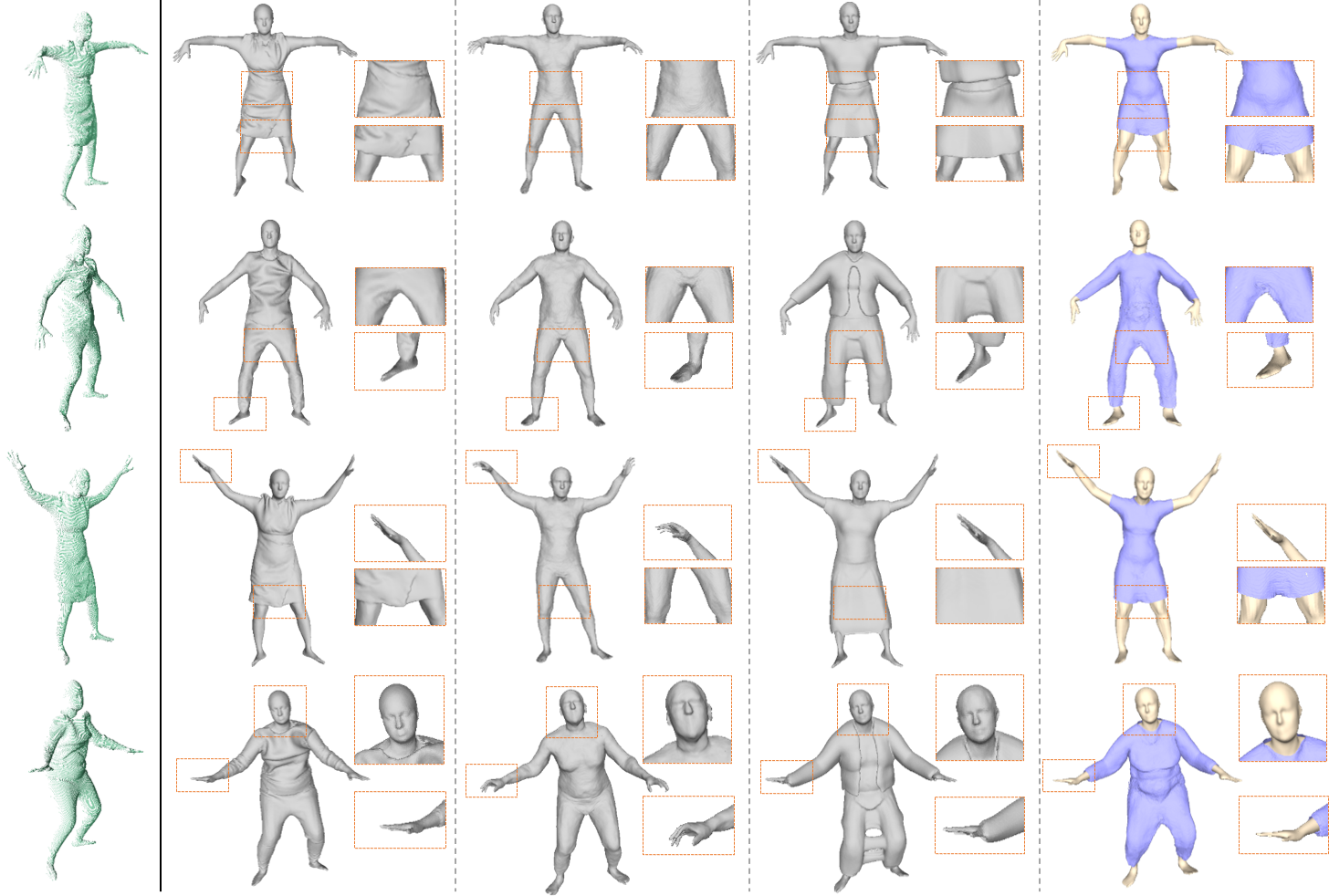}
  \begin{picture}(0,0)
    \put(-480,-9){Input}
    \put(-400,-9){GT Mesh}
    \put(-290,-9){NPMs\cite{palafox2021npms}}
    \put(-185,-9){SMPLicit\cite{corona2021smplicit}}
    \put(-70,-9){Ours}
  \end{picture}
  \vspace{0.2mm}
  \caption{Comparison to state-of-the-art methods on the task of fitting depth maps.
  }
\label{fit_depth}
\end{figure*}
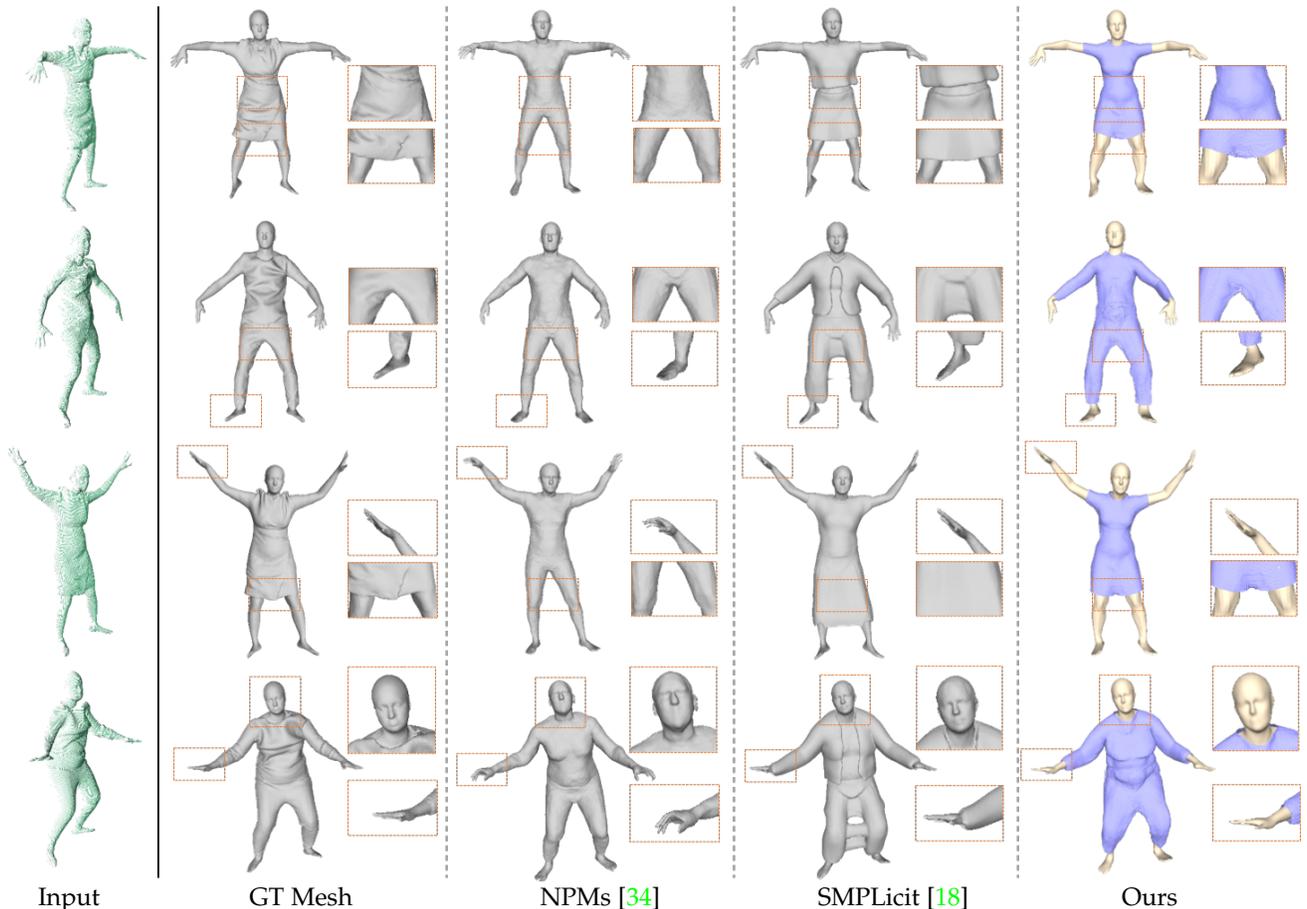

In Figure~\ref{fit_depth}, we show the comparison results of fitted monocular depth maps. 
We show the results of fitting persons in different types of garments. 
The results of NPMs are similar to tights and the dress cannot be reconstructed.
SMPLicit can reconstruct similar classes of clothes but does not fit the shapes well, since it presets the class of clothing, which limits the representative ability of the model.
Our model can achieve high-quality reconstructions from the partial inputs, and we can efficiently reconstruct the correct clothed bodies, such as wearing dress or a T-shirt.
We also show the quantitative results in Table~\ref{tab_depth}. We use IoU and chamfer distances to evaluate the reconstruction results. 
For SMPLicit, we report the IoU values after merging the SMPLicit layers, as it reconstructs the garment into separated surfaces.
Our Neural-ABC can reconstruct correct clothing categories, fit the shapes well, 
and perform best due to the representation ability on the clothed human body.

\begin{table}
\centering
\caption{Quantitative evaluations on NPMs\cite{palafox2021npms}, SMPLicit\cite{corona2021smplicit} and our Neural-ABC. 
We report the mean values of IoU and chamfer.
}  
\begin{tabular}{cccc}
\toprule   
 Method & IoU $\uparrow$ & {Chamfer}($\times10^{-2})$ $\downarrow$ \\ 
\midrule 
NPMs & 0.708 & 1.63 \\
SMPLicit & 0.636 & 1.77 \\
\textbf{Ours} & \textbf{0.742} & \textbf{1.23} \\ 
\bottomrule 
\end{tabular}
\label{tab_depth} 
\end{table}
  
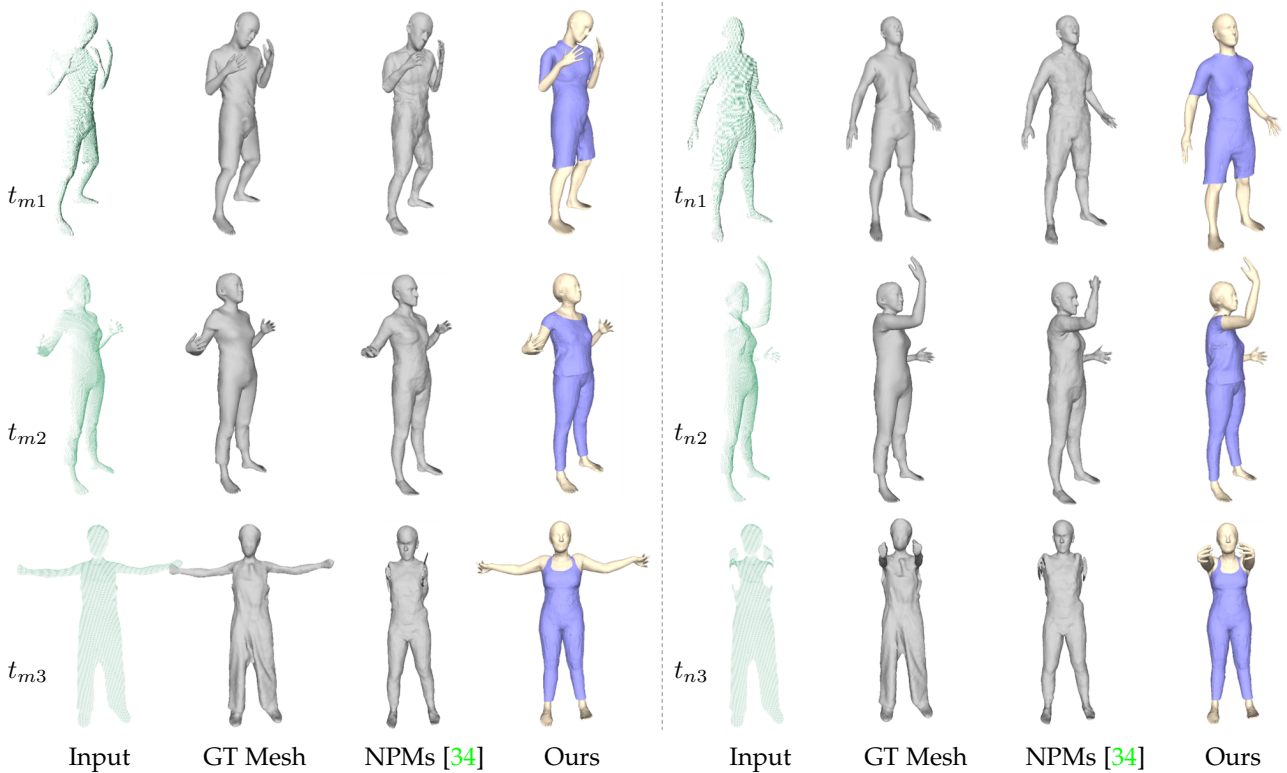
\begin{figure*}[ht]
  \centering
  \includegraphics[scale=0.32]{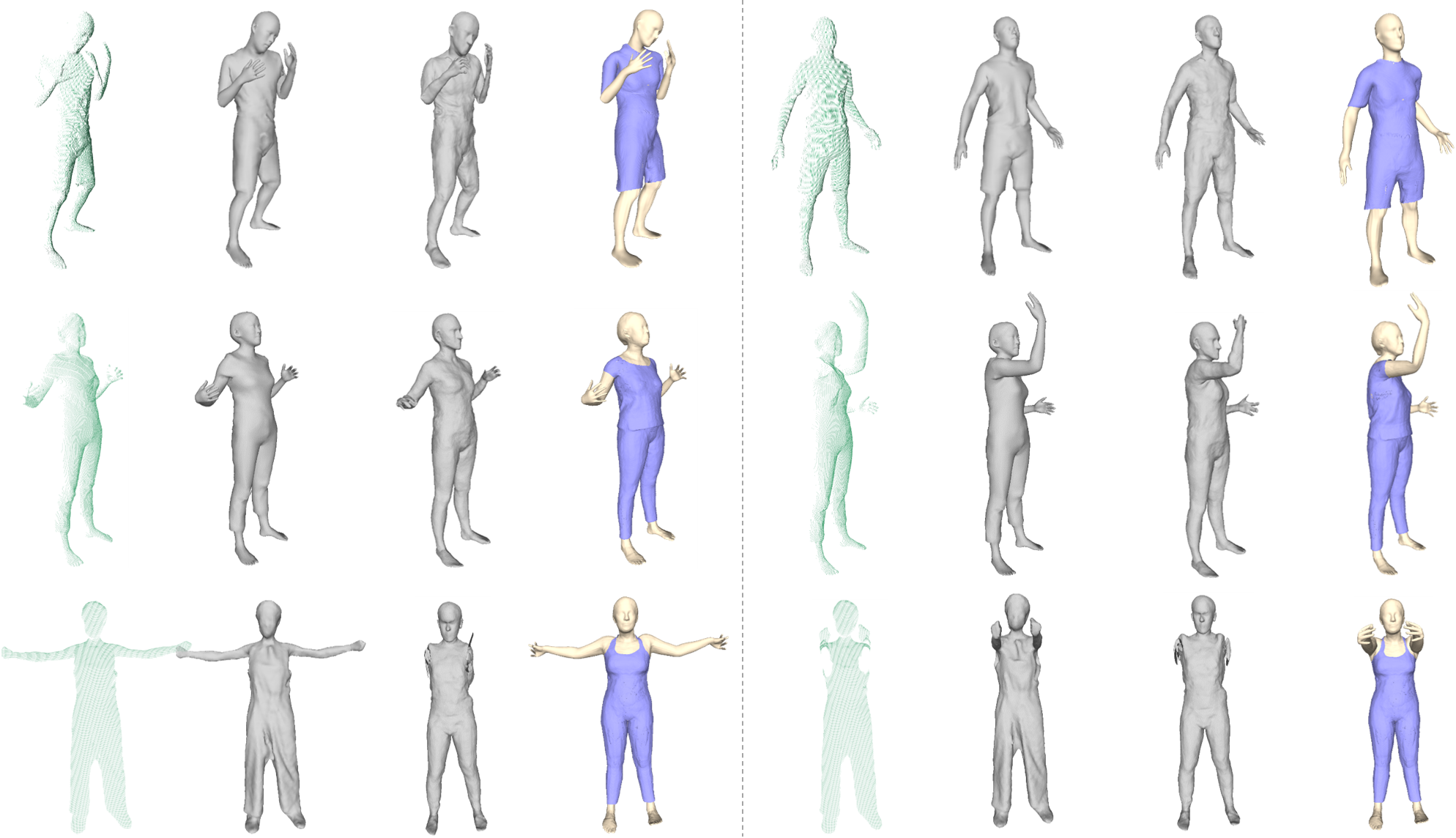}
  \begin{picture}(0,0)
    \put(-485,200){$t_{m1}$}
    \put(-485,110){$t_{m2}$}
    \put(-485,20){$t_{m3}$}
    
    \put(-462,-13){Input}
    \put(-411,-13){GT Mesh}
    \put(-350,-13){NPMs\cite{palafox2021npms}}
    \put(-282,-13){Ours}

    \put(-233,200){$t_{n1}$}
    \put(-233,110){$t_{n2}$}
    \put(-233,20){$t_{n3}$}

    \put(-212,-13){Input}
    \put(-161,-13){GT Mesh}
    \put(-100,-13){NPMs\cite{palafox2021npms}}
    \put(-32,-13){Ours}
  \end{picture}
  \vspace{3mm}
  \caption{Comparison to state-of-the-art methods on the task of fitting depth sequences.
  }
\label{depth_seq2}
\end{figure*}

We also evaluate real sequences from the CAPE\cite{ma2020learning} and Humman\cite{cai2022humman} datasets in comparsion with NPMs\cite{palafox2021npms}.
For each sequence, share the same identity and clothing code during optimization.
We show the fitting results in Figure~\ref{depth_seq2}.
NPMs were able to reconstruct tight garments, but failed to reconstruct loose garments.
We show the quantitative results in Table~\ref{tab_depth_seq}. 
Our Neural-ABC can achieve better reconstruction results for different clothing types and hairstyles.

\begin{table}
  \centering
  \caption{Quantitative evaluations on  NPMs\cite{palafox2021npms} and our Neural-ABC. 
  }  
  \begin{tabular}{cccc}
  \toprule 
   Method & IoU $\uparrow$ & {Chamfer}($\times10^{-2})$ $\downarrow$ \\ 
  \midrule 
  NPMs & 0.606 & 2.46 \\
  \textbf{Ours} & \textbf{0.762} & \textbf{1.42} \\ 
  \bottomrule 
  \end{tabular}
  \label{tab_depth_seq} 
\end{table}

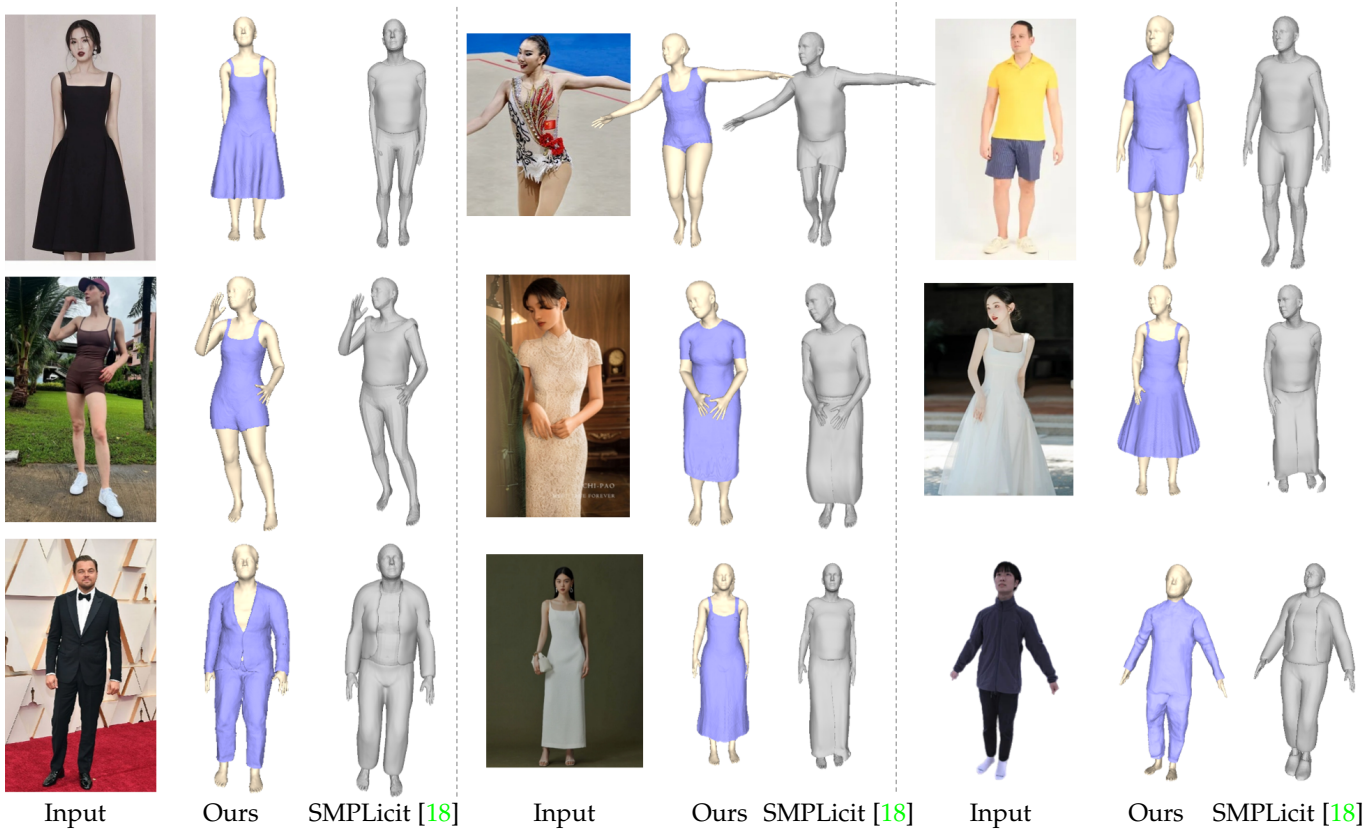
\begin{figure*}[ht]
  \centering
  \includegraphics[scale=0.35]{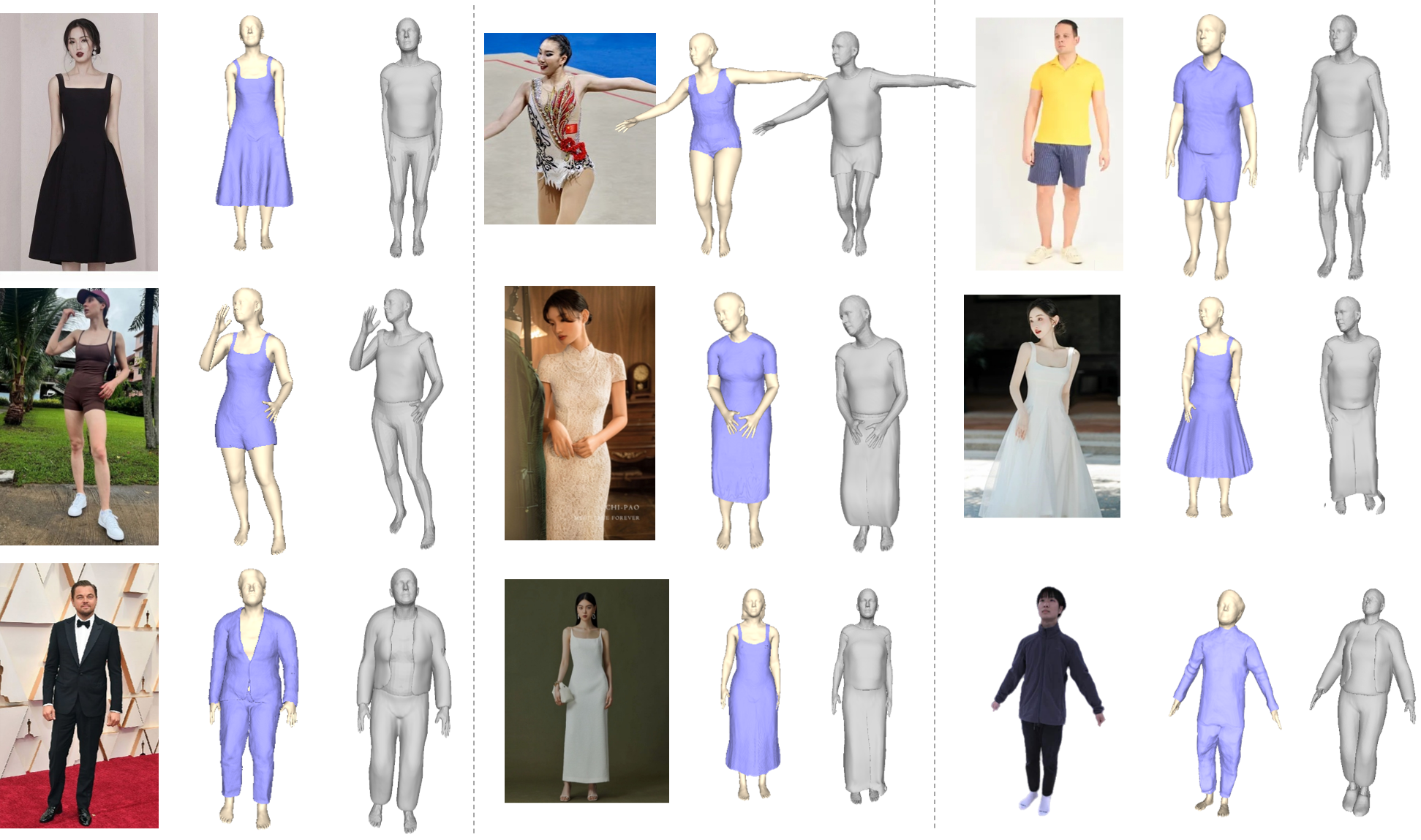}
  \begin{picture}(0,0)
      \put(-500,-7){Input}
      \put(-440,-7){Ours}
      \put(-400,-7){SMPLicit\cite{corona2021smplicit}}
      \put(-315,-7){Input}
      \put(-255,-7){Ours}
      \put(-228,-7){SMPLicit\cite{corona2021smplicit}}
      \put(-150,-7){Input}
      \put(-90,-7){Ours}
      \put(-58,-7){SMPLicit\cite{corona2021smplicit}}
  \end{picture}
  \caption{Recovering the clothed body geometry from the image. For each group, the left shows the input image, the middle shows the result of our method, and the right side shows the result of SMPLicit\cite{corona2021smplicit}.}
  \label{img_fit}
\end{figure*}

\textbf{Recovering from Images.}
We demonstrate the capability of Neural-ABC in recovering the geometry of clothed human bodies from images, in Figure~\ref{img_fit}.
We follow \cite{de2023drapenet} and extract the latent codes from the reference images. 
We compare our experimental results with SMPLicit\cite{corona2021smplicit}. 
Since SMPLicit differentiates types of clothing, for the parsing results\cite{li2020self} of "Dress", we used the "Upper-clothes" and "Skirt" models respectively, and for "Jumpsuits", we used the "Upper-clothes" and "Pants" models respectively. Our model achieves better results for different types of clothing.

\begin{figure*}[ht]
  \centering
  \includegraphics[scale=0.26]{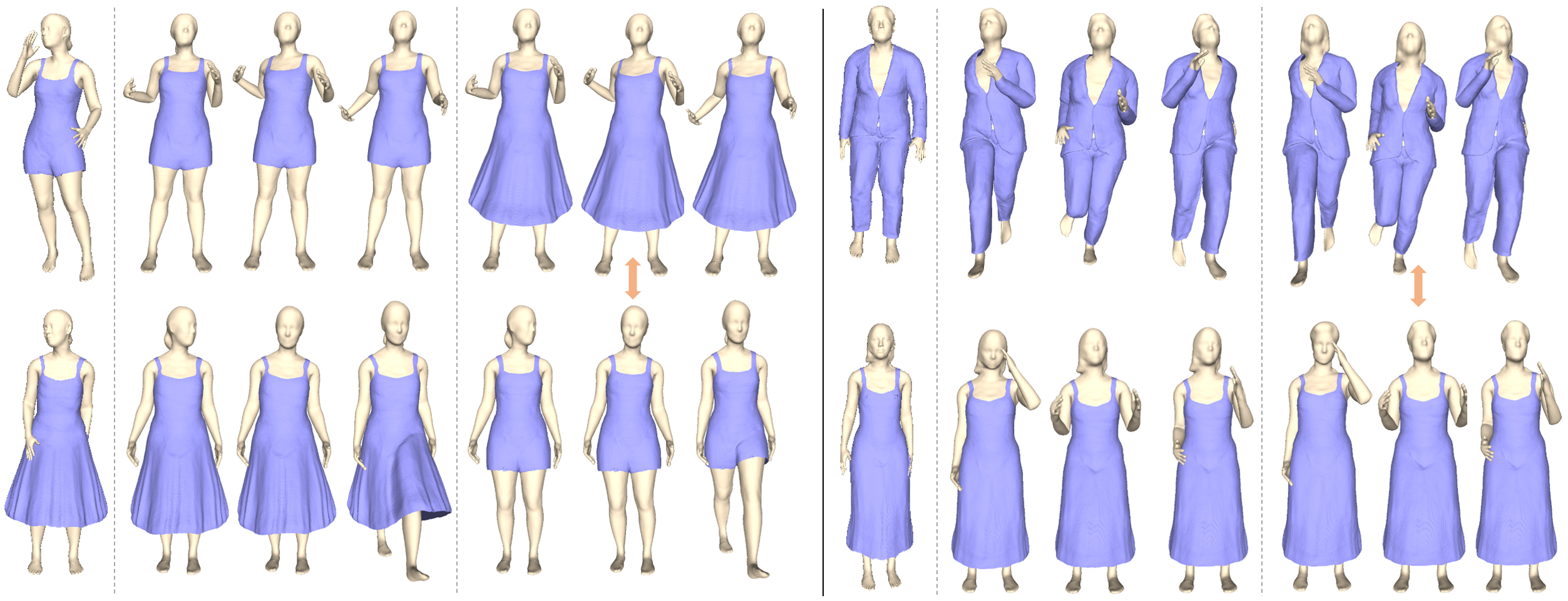}
  \begin{picture}(0,0)
      \put(-490,-10){result}
      \put(-430,-10){Animation}
      \put(-335,-10){Swapping clothing}
      \put(-230,-10){result}
      \put(-175,-10){Animation}
      \put(-90,-10){Swapping identities}
  \end{picture}
  \vspace{2mm}
  \caption{Editing the fitting results. 
    }
  \label{img_pose}
\end{figure*}

\textbf{Editing the Results.}
Editing the latent code of the fitting results can modify the corresponding geometric attributes. In Figure~\ref{img_pose}, we showcase the animation results of the fitting, including the animation results of swapping clothing and identities.
For animation, we directly modify the pose code to change the generated results. For multiple fittings, we
demonstrate the animation results of swapping clothing and identities.

\begin{figure*}[ht]
  \centering
  \includegraphics[scale=0.31]{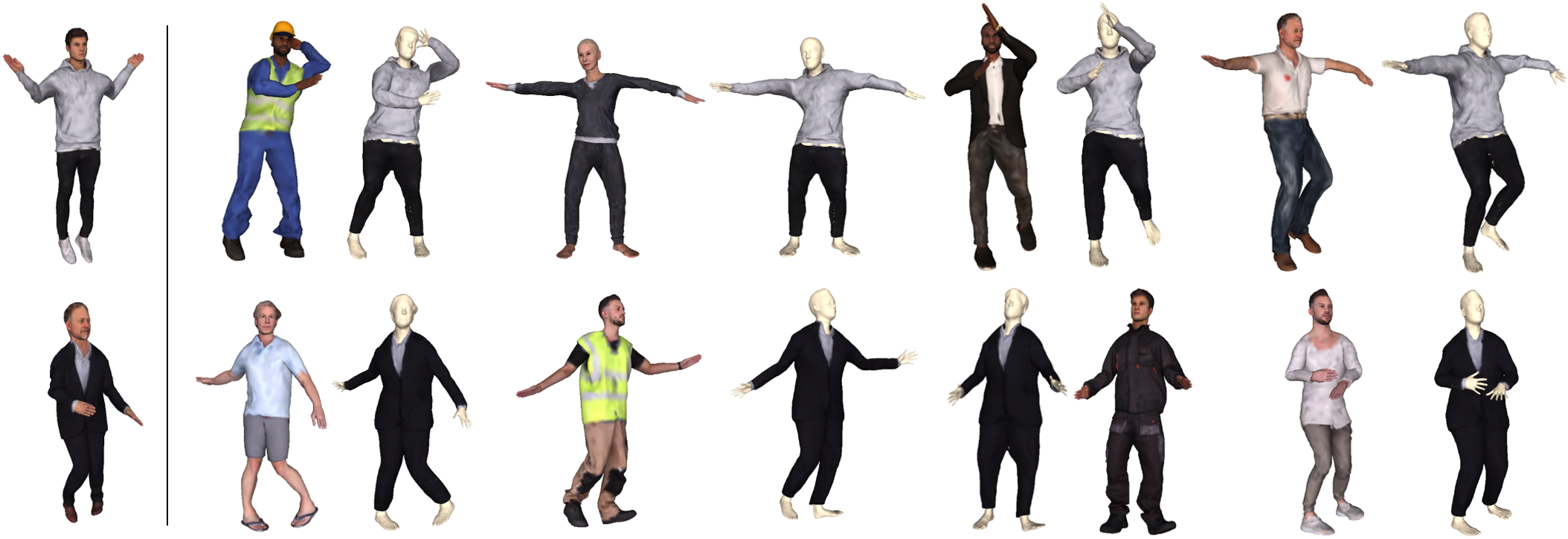}
  \begin{picture}(0,0)
      \put(-480,-10){target}
      \put(-415,-10){source}
      \put(-370,-10){result}
      \put(-310,-10){source}
      \put(-250,-10){result}
      \put(-190,-10){source}
      \put(-150,-10){result}
      \put(-85,-10){source}
      \put(-40,-10){result}
  \end{picture}
  \vspace{2mm}
  \caption{Clothing transfer results on test dataset. The target clothed bodies are shown in the first column, 
    and the clothing in the target scans is transfered to four different source clothed body scans.
    For each result, the source body with transfer clothing are presented.
    }
  \label{clothing_transfer}
\end{figure*}

\textbf{Clothing Transfer.}
Due to the independent decoupling ability of our model, clothing transfer can be achieved by exchanging clothing codes after fitting two different inputs.
We show the results of clothing transfer in Figure~\ref{clothing_transfer}.
For each set of targets and sources, we first optimize the latent codes and fit the model to the input.
Then, the clothing code of the target and the identity, shape and pose codes of the source are used to generate the results of the clothing transfer.
The vertex colors of clothing are obtained from target scans through KDTree when fitting the target for display.
Our model performs well in representing independent hoodies and suits.

\begin{figure}[ht]
  \centering
  \includegraphics[scale=0.45]{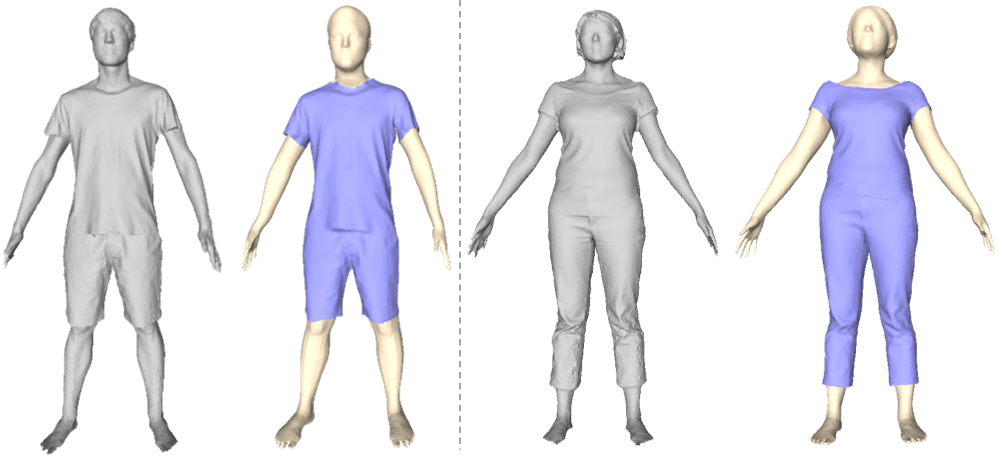}
  \begin{picture}(0,0)
      \put(-208,-10){target}
      \put(-155,-10){result}
      \put(-101,-10){target}
      \put(-41,-10){result}
  \end{picture}
  \vspace{2mm}
  \caption{Non-rigid registration applied to the fitting results.
    }
  \label{recon}
\end{figure}

\textbf{Deformation Refinement.}
Our method can be used for geometric initialization in reconstruction.
The dressed human body geometry obtained from our model can be enhanced with finer clothing details by incorporating additional deformation modules.
For the reconstruction results, we can employ non-rigid registration to obtain finer clothing wrinkles.
We demonstrate the results of deformation refinement in Figure~\ref{recon}.

\textbf{Physics-based Animation.}
For physics-based animation, we use fitting to obtain decoupled clothing and body, and then employ physics-based methods to generate animation results, achieving a more realistic deformation of the garments. In contrast, Section~\ref{Animation} Animation only produces animation results in a parameterized space.
We showcase the results in Figure~\ref{deform}, for more animated effects, please refer to the attached video.
\begin{figure}[ht]
  \centering
  \includegraphics[scale=0.52]{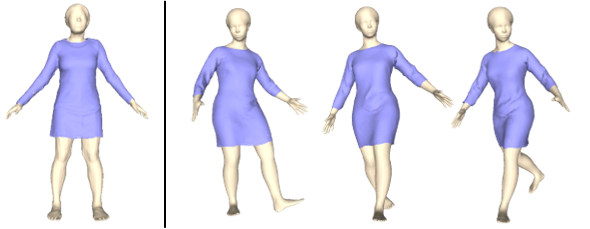}
  \begin{picture}(0,0)
      \put(-210,-10){(a)}
      \put(-85,-10){(b)}
  \end{picture}
  \vspace{2mm}
  \caption{(a) Non-rigid registration applied to the fitting results.
    (b) Clothing human body sequence driven by physics-based simulation\cite{grigorev2022hood}.
    }
  \label{deform}
\end{figure}

\section{Limitations}
Our representation works well for clothed bodies with different topologies, and is especially suitable for the representation of various styles of clothes. 
However, it may fail for some special clothes, since our training dataset does not contain all clothing types.
This can improve the representation ability of the model by collecting more training data.

The geometry represented by UDF is not limited to closed surfaces, which can represent dressed human bodies with complex topological structures.
However, there are limitations to the method of extracting implicit UDFs into meshes. 
Extracting with the Marching Cubes\cite{lorensen1987marching} yields a thin layer with thickness, which is not a suitable representation for the dressed human body. 
MeshUDF\cite{guillard2022udf} can accurately extract the zero level set, but it cannot extract nonmanifold surfaces. 
This may result in artifacts in the extracted mesh at locations where the clothes come into contact with the underlying body, or when the clothes come with some decorations (such as pockets, and buttons).
We removed decorations from the training data to avoid generating nonmanifold surfaces, and extracted clothing and underlying bodies separately during display to avoid artifacts.
In the future, this can be compensated for by detecting and processing such contact surfaces.

A parameterized model employs several low-dimensional vectors to represent various attributes of a category of objects. Due to the limited number of parameters and low dimensions, capturing certain high-frequency details of an object solely within the parameterized space, such as intricate clothing folds, may be ineffective. On the other hand, our parameterized model can be employed to provide geometric initialization for the clothed human body. If more detailed information is required, further refinement is also possible.

Modeling fine clothing wrinkles in a parametric space is a challenging problem.
Realistic clothing wrinkles often require consideration of multiple factors, such as accurate material properties, collision handling, and environmental elements like wind.
Data-driven approaches depend on a large amount of training data. However, due to the complex variations in clothing wrinkles, the generalization performance of these methods is often quite limited, leading to less realistic wrinkles in the generated results.
Physics-based simulation methods achieve good results in animations of individual examples, primarily because they can accurately model physical properties such as elasticity and inertia, as well as the state of the preceding frames.
We propose a parametric model for a dressed human body that focuses on the plausibility of posture-dependent clothing deformation, ensuring it can follow bodily changes without collision in specified poses. In our future work, we plan to consider inter-frame information in animations to develop a model capable of incorporating realistic wrinkle details in animation generation.

\section{Conclusion}
We have proposed Neural-ABC, a learned neural implicit parametric model for clothed human bodies, with disentangled identity, clothing, shape and pose spaces that can be independently controlled and edited. 
We use a unified implicit representation to model the clothing and underlying body, and use a cascading architecture to decouple the four attributes independently.
Thanks to the novel framework, Neural-ABC can represent different clothing types of posed clothed bodies without predefined templates or clothing classification.
The representation ability of Neural-ABC has been verified by fitting clothed human bodies from raw scans, depth maps and images. 
The results can be edited by latent space transfer and interpolation.
Our proposed Neural-ABC lays an important foundation for digital humans, and the separated clothing and controllable pose representation bring many possibilities for applications such as virtual try-on.

\ifCLASSOPTIONcompsoc
  \section*{Acknowledgments}
\else
  \section*{Acknowledgment}
\fi

This research was partially supported by the National Natural Science Foundation of China (No.62122071, No.62272433), and the Fundamental Research Funds for the Central Universities (No. WK3470000021).

\bibliographystyle{IEEEtran}
\bibliography{ref}

\begin{thebibliography}{10}
\providecommand{\url}[1]{#1}
\csname url@samestyle\endcsname
\providecommand{\newblock}{\relax}
\providecommand{\bibinfo}[2]{#2}
\providecommand{\BIBentrySTDinterwordspacing}{\spaceskip=0pt\relax}
\providecommand{\BIBentryALTinterwordstretchfactor}{4}
\providecommand{\BIBentryALTinterwordspacing}{\spaceskip=\fontdimen2\font plus
\BIBentryALTinterwordstretchfactor\fontdimen3\font minus \fontdimen4\font\relax}
\providecommand{\BIBforeignlanguage}[2]{{%
\expandafter\ifx\csname l@#1\endcsname\relax
\typeout{** WARNING: IEEEtran.bst: No hyphenation pattern has been}%
\typeout{** loaded for the language `#1'. Using the pattern for}%
\typeout{** the default language instead.}%
\else
\language=\csname l@#1\endcsname
\fi
#2}}
\providecommand{\BIBdecl}{\relax}
\BIBdecl

\bibitem{Kanazawa_2018_CVPR}
A.~Kanazawa, M.~J. Black, D.~W. Jacobs, and J.~Malik, ``End-to-end recovery of human shape and pose,'' in \emph{IEEE/CVF Conference on Computer Vision and Pattern Recognition (CVPR)}, June 2018.

\bibitem{plankers2001articulated}
R.~Plankers and P.~Fua, ``Articulated soft objects for video-based body modeling,'' in \emph{Proceedings Eighth IEEE International Conference on Computer Vision. ICCV 2001}, vol.~1.\hskip 1em plus 0.5em minus 0.4em\relax IEEE, 2001, pp. 394--401.

\bibitem{sminchisescu2006learning}
C.~Sminchisescu, A.~Kanaujia, and D.~Metaxas, ``Learning joint top-down and bottom-up processes for 3d visual inference,'' in \emph{2006 IEEE Computer Society Conference on Computer Vision and Pattern Recognition (CVPR'06)}, vol.~2.\hskip 1em plus 0.5em minus 0.4em\relax IEEE, 2006, pp. 1743--1752.

\bibitem{li2022cliff}
Z.~Li, J.~Liu, Z.~Zhang, S.~Xu, and Y.~Yan, ``Cliff: Carrying location information in full frames into human pose and shape estimation,'' in \emph{ECCV}, 2022.

\bibitem{petrovich21actor}
M.~Petrovich, M.~J. Black, and G.~Varol, ``Action-conditioned 3{D} human motion synthesis with transformer {VAE},'' in \emph{International Conference on Computer Vision (ICCV)}, 2021.

\bibitem{sminchisescu2002human}
C.~Sminchisescu and A.~Telea, ``Human pose estimation from silhouettes. a consistent approach using distance level sets,'' in \emph{International Conference on Computer Graphics, Visualization and Computer Vision (WSCG'02)}, vol.~10, 2002.

\bibitem{li2022avatarcap}
Z.~Li, Z.~Zheng, H.~Zhang, C.~Ji, and Y.~Liu, ``Avatarcap: Animatable avatar conditioned monocular human volumetric capture,'' in \emph{European Conference on Computer Vision (ECCV)}, October 2022.

\bibitem{zhi2020texmesh}
T.~Zhi, C.~Lassner, T.~Tung, C.~Stoll, S.~G. Narasimhan, and M.~Vo, ``Texmesh: Reconstructing detailed human texture and geometry from rgb-d video,'' in \emph{European Conference on Computer Vision (ECCV)}.\hskip 1em plus 0.5em minus 0.4em\relax Springer, 2020, pp. 492--509.

\bibitem{saito2021scanimate}
S.~Saito, J.~Yang, Q.~Ma, and M.~J. Black, ``Scanimate: Weakly supervised learning of skinned clothed avatar networks,'' in \emph{IEEE/CVF Conference on Computer Vision and Pattern Recognition (CVPR)}, 2021, pp. 2886--2897.

\bibitem{MetaAvatar:NeurIPS:2021}
S.~Wang, M.~Mihajlovic, Q.~Ma, A.~Geiger, and S.~Tang, ``Metaavatar: Learning animatable clothed human models from few depth images,'' in \emph{Advances in Neural Information Processing Systems}, 2021.

\bibitem{chen2022gdna}
X.~Chen, T.~Jiang, J.~Song, J.~Yang, M.~J. Black, A.~Geiger, and O.~Hilliges, ``gdna: Towards generative detailed neural avatars,'' in \emph{IEEE Conf. on Computer Vision and Pattern Recognition (CVPR)}, 2022.

\bibitem{zanfir2020human}
M.~Zanfir, E.~Oneata, A.-I. Popa, A.~Zanfir, and C.~Sminchisescu, ``Human synthesis and scene compositing,'' in \emph{AAAI Conference on Artificial Intelligence}, vol.~34, no.~07, 2020, pp. 12\,749--12\,756.

\bibitem{Anguelov05scape:shape}
D.~Anguelov, P.~Srinivasan, D.~Koller, S.~Thrun, J.~Rodgers, and J.~Davis, ``Scape: shape completion and animation of people,'' \emph{ACM Transactions on Graphics (TOG)}, vol.~24, pp. 408--416, 2005.

\bibitem{loper2015smpl}
M.~Loper, N.~Mahmood, J.~Romero, G.~Pons-Moll, and M.~J. Black, ``Smpl: A skinned multi-person linear model,'' \emph{ACM transactions on graphics (TOG)}, vol.~34, no.~6, pp. 1--16, 2015.

\bibitem{jiang2020disentangled}
B.~Jiang, J.~Zhang, J.~Cai, and J.~Zheng, ``Disentangled human body embedding based on deep hierarchical neural network,'' \emph{IEEE transactions on visualization and computer graphics}, vol.~26, no.~8, pp. 2560--2575, 2020.

\bibitem{romero2017embodied}
J.~Romero, D.~Tzionas, and M.~J. Black, ``Embodied hands: Modeling and capturing hands and bodies together,'' \emph{ACM Transactions on Graphics (TOG)}, vol.~36, no.~6, pp. 1--17, 2017.

\bibitem{jiang2020bcnet}
B.~Jiang, J.~Zhang, Y.~Hong, J.~Luo, L.~Liu, and H.~Bao, ``Bcnet: Learning body and cloth shape from a single image,'' in \emph{European Conference on Computer Vision (ECCV)}.\hskip 1em plus 0.5em minus 0.4em\relax Springer, 2020, pp. 18--35.

\bibitem{corona2021smplicit}
E.~Corona, A.~Pumarola, G.~Alenya, G.~Pons-Moll, and F.~Moreno-Noguer, ``Smplicit: Topology-aware generative model for clothed people,'' in \emph{IEEE/CVF Conference on Computer Vision and Pattern Recognition (CVPR)}, 2021, pp. 11\,875--11\,885.

\bibitem{ren2022dig}
L.~Ren, B.~Guillard, E.~Remelli, and P.~Fua, ``{DIG: Draping Implicit Garment over the Human Body},'' in \emph{Asian Conference on Computer Vision}, 2022.

\bibitem{de2023drapenet}
L.~De~Luigi, R.~Li, B.~Guillard, M.~Salzmann, and P.~Fua, ``{DrapeNet: Garment Generation and Self-Supervised Draping},'' in \emph{Proceedings of the IEEE/CVF Conference on Computer Vision and Pattern Recognition}, 2023.

\bibitem{pons2017clothcap}
G.~Pons-Moll, S.~Pujades, S.~Hu, and M.~J. Black, ``Clothcap: Seamless 4d clothing capture and retargeting,'' \emph{ACM Transactions on Graphics (TOG)}, vol.~36, no.~4, pp. 1--15, 2017.

\bibitem{zhang2017detailed}
C.~Zhang, S.~Pujades, M.~J. Black, and G.~Pons-Moll, ``Detailed, accurate, human shape estimation from clothed 3d scan sequences,'' in \emph{Proceedings of the IEEE Conference on Computer Vision and Pattern Recognition}, 2017, pp. 4191--4200.

\bibitem{alldieck2018video}
T.~Alldieck, M.~Magnor, W.~Xu, C.~Theobalt, and G.~Pons-Moll, ``Video based reconstruction of 3d people models,'' in \emph{{IEEE} Conference on Computer Vision and Pattern Recognition}, 2018.

\bibitem{ThiemoAlldieck2019Tex2ShapeDF}
T.~Alldieck, G.~Pons-Moll, C.~Theobalt, and M.~Magnor, ``Tex2shape: Detailed full human body geometry from a single image,'' \emph{International Conference on Computer Vision}, 2019.

\bibitem{alldieck2019learning}
T.~Alldieck, M.~Magnor, B.~L. Bhatnagar, C.~Theobalt, and G.~Pons-Moll, ``Learning to reconstruct people in clothing from a single rgb camera,'' in \emph{IEEE/CVF Conference on Computer Vision and Pattern Recognition (CVPR)}, 2019, pp. 1175--1186.

\bibitem{bhatnagar2019mgn}
B.~L. Bhatnagar, G.~Tiwari, C.~Theobalt, and G.~Pons-Moll, ``Multi-garment net: Learning to dress 3d people from images,'' in \emph{{IEEE} International Conference on Computer Vision ({ICCV})}.\hskip 1em plus 0.5em minus 0.4em\relax {IEEE}, Oct 2019.

\bibitem{ma2020learning}
Q.~Ma, J.~Yang, A.~Ranjan, S.~Pujades, G.~Pons-Moll, S.~Tang, and M.~J. Black, ``Learning to dress 3d people in generative clothing,'' in \emph{IEEE/CVF Conference on Computer Vision and Pattern Recognition (CVPR)}, 2020, pp. 6469--6478.

\bibitem{bhatnagar2020loopreg}
B.~L. Bhatnagar, C.~Sminchisescu, C.~Theobalt, and G.~Pons-Moll, ``Loopreg: Self-supervised learning of implicit surface correspondences, pose and shape for 3d human mesh registration,'' in \emph{Neural Information Processing Systems (NeurIPS)}, December 2020.

\bibitem{deng2020nasa}
B.~Deng, J.~P. Lewis, T.~Jeruzalski, G.~Pons-Moll, G.~Hinton, M.~Norouzi, and A.~Tagliasacchi, ``Nasa neural articulated shape approximation,'' in \emph{European Conference on Computer Vision (ECCV)}.\hskip 1em plus 0.5em minus 0.4em\relax Springer, 2020, pp. 612--628.

\bibitem{chen2021snarf}
X.~Chen, Y.~Zheng, M.~J. Black, O.~Hilliges, and A.~Geiger, ``Snarf: Differentiable forward skinning for animating non-rigid neural implicit shapes,'' in \emph{International Conference on Computer Vision (ICCV)}, 2021.

\bibitem{tiwari2021neural}
G.~Tiwari, N.~Sarafianos, T.~Tung, and G.~Pons-Moll, ``Neural-gif: Neural generalized implicit functions for animating people in clothing,'' in \emph{IEEE/CVF International Conference on Computer Vision (ICCV)}, 2021, pp. 11\,708--11\,718.

\bibitem{mihajlovic2021leap}
M.~Mihajlovic, Y.~Zhang, M.~J. Black, and S.~Tang, ``Leap: Learning articulated occupancy of people,'' in \emph{IEEE/CVF Conference on Computer Vision and Pattern Recognition (CVPR)}, 2021, pp. 10\,461--10\,471.

\bibitem{alldieck2021imghum}
T.~Alldieck, H.~Xu, and C.~Sminchisescu, ``imghum: Implicit generative models of 3d human shape and articulated pose,'' in \emph{IEEE/CVF International Conference on Computer Vision (ICCV)}, 2021, pp. 5461--5470.

\bibitem{palafox2021npms}
P.~Palafox, A.~Bo{\v{z}}i{\v{c}}, J.~Thies, M.~Nie{\ss}ner, and A.~Dai, ``Npms: Neural parametric models for 3d deformable shapes,'' in \emph{IEEE/CVF International Conference on Computer Vision (ICCV)}, 2021.

\bibitem{yu2019simulcap}
T.~Yu, Z.~Zheng, Y.~Zhong, J.~Zhao, Q.~Dai, G.~Pons-Moll, and Y.~Liu, ``Simulcap: Single-view human performance capture with cloth simulation,'' in \emph{Proceedings of the IEEE/CVF conference on computer vision and pattern recognition}, 2019, pp. 5504--5514.

\bibitem{Feng2022scarf}
Y.~Feng, J.~Yang, M.~Pollefeys, M.~J. Black, and T.~Bolkart, ``Capturing and animation of body and clothing from monocular video,'' in \emph{SIGGRAPH Asia 2022 Conference Papers}, ser. SA '22, 2022.

\bibitem{guillard2022udf}
B.~Guillard, F.~Stella, and P.~Fua, ``Meshudf: Fast and differentiable meshing of unsigned distance field networks,'' in \emph{European Conference on Computer Vision}, 2022.

\bibitem{zhao2021learning}
F.~Zhao, W.~Wang, S.~Liao, and L.~Shao, ``Learning anchored unsigned distance functions with gradient direction alignment for single-view garment reconstruction,'' in \emph{Proceedings of the IEEE/CVF International Conference on Computer Vision}, 2021, pp. 12\,674--12\,683.

\bibitem{long2022neuraludf}
X.~Long, C.~Lin, L.~Liu, Y.~Liu, P.~Wang, C.~Theobalt, T.~Komura, and W.~Wang, ``Neuraludf: Learning unsigned distance fields for multi-view reconstruction of surfaces with arbitrary topologies,'' \emph{arXiv preprint arXiv:2211.14173}, 2022.

\bibitem{chibane2020ndf}
J.~Chibane, A.~Mir, and G.~Pons-Moll, ``Neural unsigned distance fields for implicit function learning,'' in \emph{Advances in Neural Information Processing Systems ({NeurIPS})}, December 2020.

\bibitem{xu2020ghum}
H.~Xu, E.~G. Bazavan, A.~Zanfir, W.~T. Freeman, R.~Sukthankar, and C.~Sminchisescu, ``Ghum \& ghuml: Generative 3d human shape and articulated pose models,'' in \emph{IEEE/CVF Conference on Computer Vision and Pattern Recognition (CVPR)}, 2020, pp. 6184--6193.

\bibitem{pavlakos2019expressive}
G.~Pavlakos, V.~Choutas, N.~Ghorbani, T.~Bolkart, A.~A. Osman, D.~Tzionas, and M.~J. Black, ``Expressive body capture: 3d hands, face, and body from a single image,'' in \emph{IEEE/CVF Conference on Computer Vision and Pattern Recognition (CVPR)}, 2019, pp. 10\,975--10\,985.

\bibitem{hesse2019learning}
N.~Hesse, S.~Pujades, M.~J. Black, M.~Arens, U.~G. Hofmann, and A.~S. Schroeder, ``Learning and tracking the 3d body shape of freely moving infants from rgb-d sequences,'' \emph{IEEE Transactions on Pattern Analysis and Machine Intelligence}, vol.~42, no.~10, pp. 2540--2551, 2019.

\bibitem{osman2020star}
A.~A. Osman, T.~Bolkart, and M.~J. Black, ``Star: Sparse trained articulated human body regressor,'' in \emph{European Conference on Computer Vision (ECCV)}.\hskip 1em plus 0.5em minus 0.4em\relax Springer, 2020, pp. 598--613.

\bibitem{joo2018total}
H.~Joo, T.~Simon, and Y.~Sheikh, ``Total capture: A 3d deformation model for tracking faces, hands, and bodies,'' in \emph{IEEE Conference on Computer Vision and Pattern Recognition (CVPR)}, 2018, pp. 8320--8329.

\bibitem{mahmood2019amass}
N.~Mahmood, N.~Ghorbani, N.~F. Troje, G.~Pons-Moll, and M.~J. Black, ``Amass: Archive of motion capture as surface shapes,'' in \emph{IEEE/CVF International Conference on Computer Vision (ICCV)}, 2019, pp. 5442--5451.

\bibitem{santesteban2020softsmpl}
I.~Santesteban, E.~Garces, M.~A. Otaduy, and D.~Casas, ``Softsmpl: Data-driven modeling of nonlinear soft-tissue dynamics for parametric humans,'' in \emph{Computer Graphics Forum}, vol.~39, no.~2.\hskip 1em plus 0.5em minus 0.4em\relax Wiley Online Library, 2020, pp. 65--75.

\bibitem{deepcloth_su2022}
Z.~Su, T.~Yu, Y.~Wang, and Y.~Liu, ``Deepcloth: Neural garment representation for shape and style editing,'' \emph{IEEE Transactions on Pattern Analysis and Machine Intelligence}, vol.~45, no.~2, pp. 1581--1593, 2023.

\bibitem{santesteban2021garmentcollisions}
I.~Santesteban, N.~Thuerey, M.~A. Otaduy, and D.~Casas, ``{Self-Supervised Collision Handling via Generative 3D Garment Models for Virtual Try-On},'' \emph{IEEE/CVF Conference on Computer Vision and Pattern Recognition (CVPR)}, 2021.

\bibitem{yang2018physics}
S.~Yang, Z.~Pan, T.~Amert, K.~Wang, L.~Yu, T.~Berg, and M.~C. Lin, ``Physics-inspired garment recovery from a single-view image,'' \emph{ACM Transactions on Graphics (TOG)}, vol.~37, no.~5, pp. 1--14, 2018.

\bibitem{su2020mulaycap}
Z.~Su, W.~Wan, T.~Yu, L.~Liu, L.~Fang, W.~Wang, and Y.~Liu, ``Mulaycap: Multi-layer human performance capture using a monocular video camera,'' 2020.

\bibitem{lorensen1987marching}
W.~E. Lorensen and H.~E. Cline, ``Marching cubes: A high resolution 3d surface construction algorithm,'' \emph{ACM siggraph computer graphics}, vol.~21, no.~4, pp. 163--169, 1987.

\bibitem{saito2019pifu}
S.~Saito, Z.~Huang, R.~Natsume, S.~Morishima, A.~Kanazawa, and H.~Li, ``Pifu: Pixel-aligned implicit function for high-resolution clothed human digitization,'' in \emph{IEEE/CVF International Conference on Computer Vision (ICCV)}, 2019, pp. 2304--2314.

\bibitem{he2020geopifu}
T.~He, J.~Collomosse, H.~Jin, and S.~Soatto, ``Geo-pifu: Geometry and pixel aligned implicit functions for single-view human reconstruction,'' in \emph{Conference on Neural Information Processing Systems (NeurIPS)}, 2020.

\bibitem{saito2020pifuhd}
S.~Saito, T.~Simon, J.~Saragih, and H.~Joo, ``Pifuhd: Multi-level pixel-aligned implicit function for high-resolution 3d human digitization,'' in \emph{IEEE Conference on Computer Vision and Pattern Recognition}, June 2020.

\bibitem{hong2021stereopifu}
Y.~Hong, J.~Zhang, B.~Jiang, Y.~Guo, L.~Liu, and H.~Bao, ``Stereopifu: Depth aware clothed human digitization via stereo vision,'' in \emph{IEEE/CVF Conference on Computer Vision and Pattern Recognition (CVPR)}, 2021, pp. 535--545.

\bibitem{zheng2021deepmulticap}
Y.~Zheng, R.~Shao, Y.~Zhang, T.~Yu, Z.~Zheng, Q.~Dai, and Y.~Liu, ``Deepmulticap: Performance capture of multiple characters using sparse multiview cameras,'' in \emph{IEEE Conference on Computer Vision (ICCV 2021)}, 2021.

\bibitem{ijcai2022p218}
\BIBentryALTinterwordspacing
G.~Yao, H.~Wu, Y.~Yuan, L.~Li, K.~Zhou, and X.~Yu, ``Learning implicit body representations from double diffusion based neural radiance fields,'' in \emph{Proceedings of the Thirty-First International Joint Conference on Artificial Intelligence, {IJCAI-22}}, L.~D. Raedt, Ed.\hskip 1em plus 0.5em minus 0.4em\relax International Joint Conferences on Artificial Intelligence Organization, 7 2022, pp. 1566--1572, main Track. [Online]. Available: \url{https://doi.org/10.24963/ijcai.2022/218}
\BIBentrySTDinterwordspacing

\bibitem{xiu2022icon}
Y.~Xiu, J.~Yang, D.~Tzionas, and M.~J. Black, ``{ICON}: {I}mplicit {C}lothed humans {O}btained from {N}ormals,'' in \emph{CVPR}, 2022.

\bibitem{huang2020arch}
Z.~Huang, Y.~Xu, C.~Lassner, H.~Li, and T.~Tung, ``Arch: Animatable reconstruction of clothed humans,'' in \emph{Proceedings of the IEEE/CVF Conference on Computer Vision and Pattern Recognition}, 2020, pp. 3093--3102.

\bibitem{chen2022fastsnarf}
X.~Chen, T.~Jiang, J.~Song, M.~Rietmann, A.~Geiger, M.~J. Black, and O.~Hilliges, ``{Fast-SNARF}: {A} fast deformer for articulated neural fields,'' \emph{arXiv}, vol. abs/2211.15601, 2022.

\bibitem{dong2022pina}
Z.~Dong, C.~Guo, J.~Song, X.~Chen, A.~Geiger, and O.~Hilliges, ``{PINA}: {L}earning a personalized implicit neural avatar from a single {RGB-D} video sequence,'' in \emph{CVPR}, 2022.

\bibitem{con_siren}
I.~Mehta, M.~Gharbi, C.~Barnes, E.~Shechtman, R.~Ramamoorthi, and M.~Chandraker, ``Modulated periodic activations for generalizable local functional representations,'' in \emph{2021 IEEE/CVF International Conference on Computer Vision (ICCV)}, 2021, pp. 14\,194--14\,203.

\bibitem{Sitzmann_Martel_Bergman_Lindell_Wetzstein_2020}
V.~Sitzmann, J.~Martel, A.~Bergman, D.~Lindell, and G.~Wetzstein, ``\BIBforeignlanguage{en-US}{Implicit neural representations with periodic activation functions},'' \emph{\BIBforeignlanguage{en-US}{Neural Information Processing Systems,Neural Information Processing Systems}}, Jun 2020.

\bibitem{park2019deepsdf}
J.~J. Park, P.~Florence, J.~Straub, R.~Newcombe, and S.~Lovegrove, ``Deepsdf: Learning continuous signed distance functions for shape representation,'' in \emph{IEEE/CVF Conference on Computer Vision and Pattern Recognition (CVPR)}, 2019, pp. 165--174.

\bibitem{icml2020_2086}
A.~Gropp, L.~Yariv, N.~Haim, M.~Atzmon, and Y.~Lipman, ``Implicit geometric regularization for learning shapes,'' in \emph{Proceedings of Machine Learning and Systems 2020}, 2020, pp. 3569--3579.

\bibitem{renderpeople}
``Website: https://renderpeople.com/,'' 2023.

\bibitem{axyz}
``Website: https://secure.axyz-design.com/,'' 2023.

\bibitem{bertiche2020cloth3d}
H.~Bertiche, M.~Madadi, and S.~Escalera, ``Cloth3d: clothed 3d humans,'' in \emph{European Conference on Computer Vision}.\hskip 1em plus 0.5em minus 0.4em\relax Springer, 2020, pp. 344--359.

\bibitem{clo3d}
``Website: https://www.clo3d.com/,'' 2023.

\bibitem{dgcnn}
Y.~Wang, Y.~Sun, Z.~Liu, S.~E. Sarma, M.~M. Bronstein, and J.~M. Solomon, ``Dynamic graph cnn for learning on point clouds,'' \emph{ACM Transactions on Graphics (TOG)}, 2019.

\bibitem{tiwari20sizer}
G.~Tiwari, B.~L. Bhatnagar, T.~Tung, and G.~Pons-Moll, ``Sizer: A dataset and model for parsing 3d clothing and learning size sensitive 3d clothing,'' in \emph{European Conference on Computer Vision ({ECCV})}.\hskip 1em plus 0.5em minus 0.4em\relax {Springer}, August 2020.

\bibitem{POP:ICCV:2021}
Q.~Ma, J.~Yang, S.~Tang, and M.~J. Black, ``The power of points for modeling humans in clothing,'' in \emph{IEEE/CVF International Conference on Computer Vision (ICCV)}, Oct. 2021.

\bibitem{kazhdan2006poisson}
M.~Kazhdan, M.~Bolitho, and H.~Hoppe, ``Poisson surface reconstruction,'' in \emph{Proceedings of the fourth Eurographics symposium on Geometry processing}, vol.~7, 2006.

\bibitem{cai2022humman}
Z.~Cai, D.~Ren, A.~Zeng, Z.~Lin, T.~Yu, W.~Wang, X.~Fan, Y.~Gao, Y.~Yu, L.~Pan, F.~Hong, M.~Zhang, C.~C. Loy, L.~Yang, and Z.~Liu, ``{HuMMan}: Multi-modal 4d human dataset for versatile sensing and modeling,'' in \emph{17th European Conference on Computer Vision, Tel Aviv, Israel, October 23--27, 2022, Proceedings, Part VII}.\hskip 1em plus 0.5em minus 0.4em\relax Springer, 2022, pp. 557--577.

\bibitem{li2020self}
P.~Li, Y.~Xu, Y.~Wei, and Y.~Yang, ``Self-correction for human parsing,'' \emph{IEEE Transactions on Pattern Analysis and Machine Intelligence}, 2020.

\bibitem{grigorev2022hood}
A.~Grigorev, B.~Thomaszewski, M.~J. Black, and O.~Hilliges, ``{HOOD}: Hierarchical graphs for generalized modelling of clothing dynamics,'' 2023.

\end{thebibliography}
\begin{IEEEbiography}[{\includegraphics[width=1in]{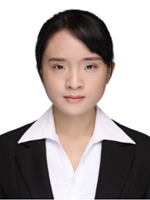}}]{Honghu Chen} is currently working toward a PhD degree in the School of Data Sciences, University of Science and Technology of China. Her research interests include computer vision, 3D vision and human modeling.
\end{IEEEbiography}
\begin{IEEEbiography}[{\includegraphics[width=1in]{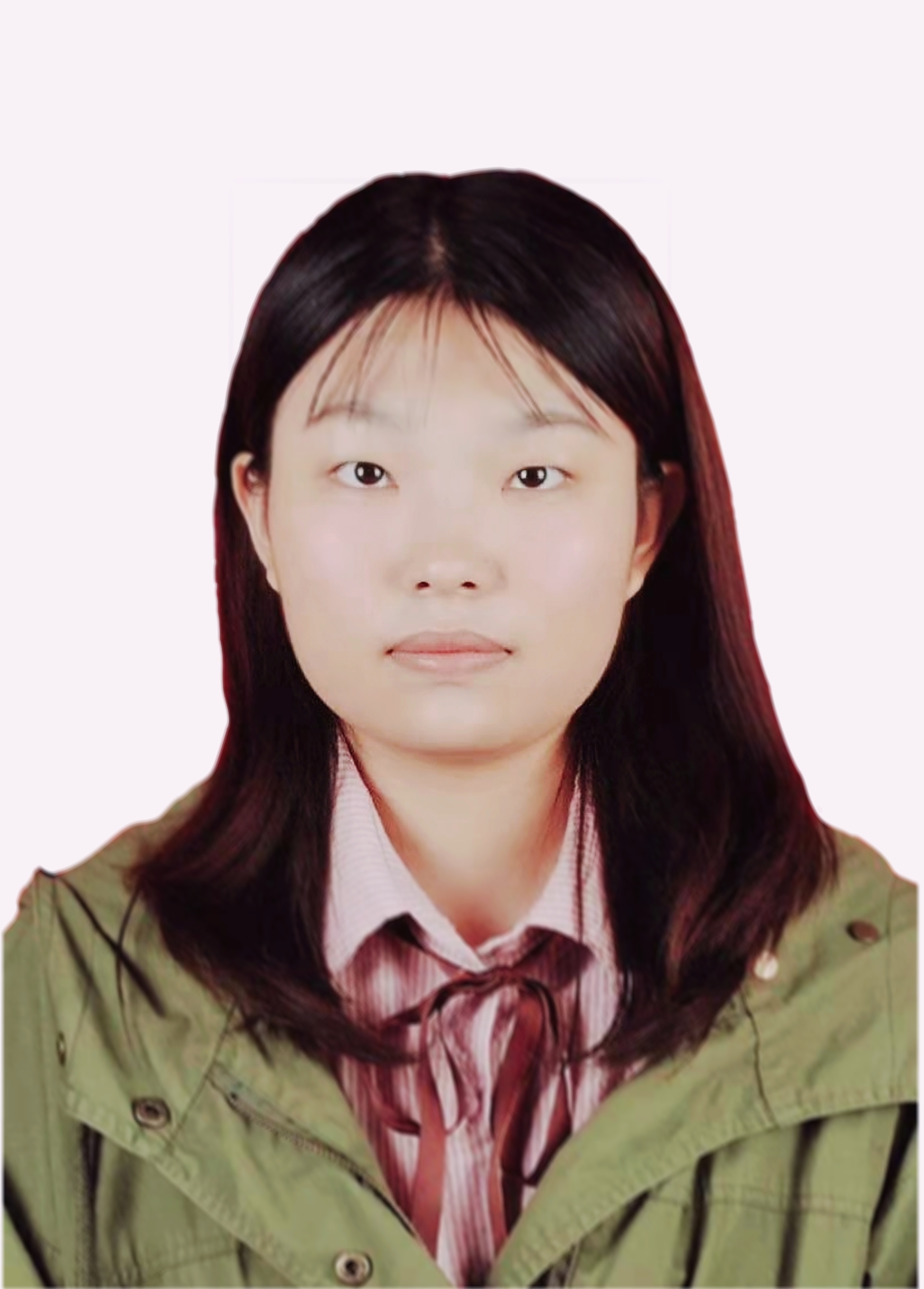}}]{Yuxin Yao} is currently working toward a PhD degree in the School of Mathematical Sciences, University of Science and Technology of China. Her research interests include computer vision, computer graphics and numerical optimization.
\end{IEEEbiography}
\begin{IEEEbiography}[{\includegraphics[width=1in]{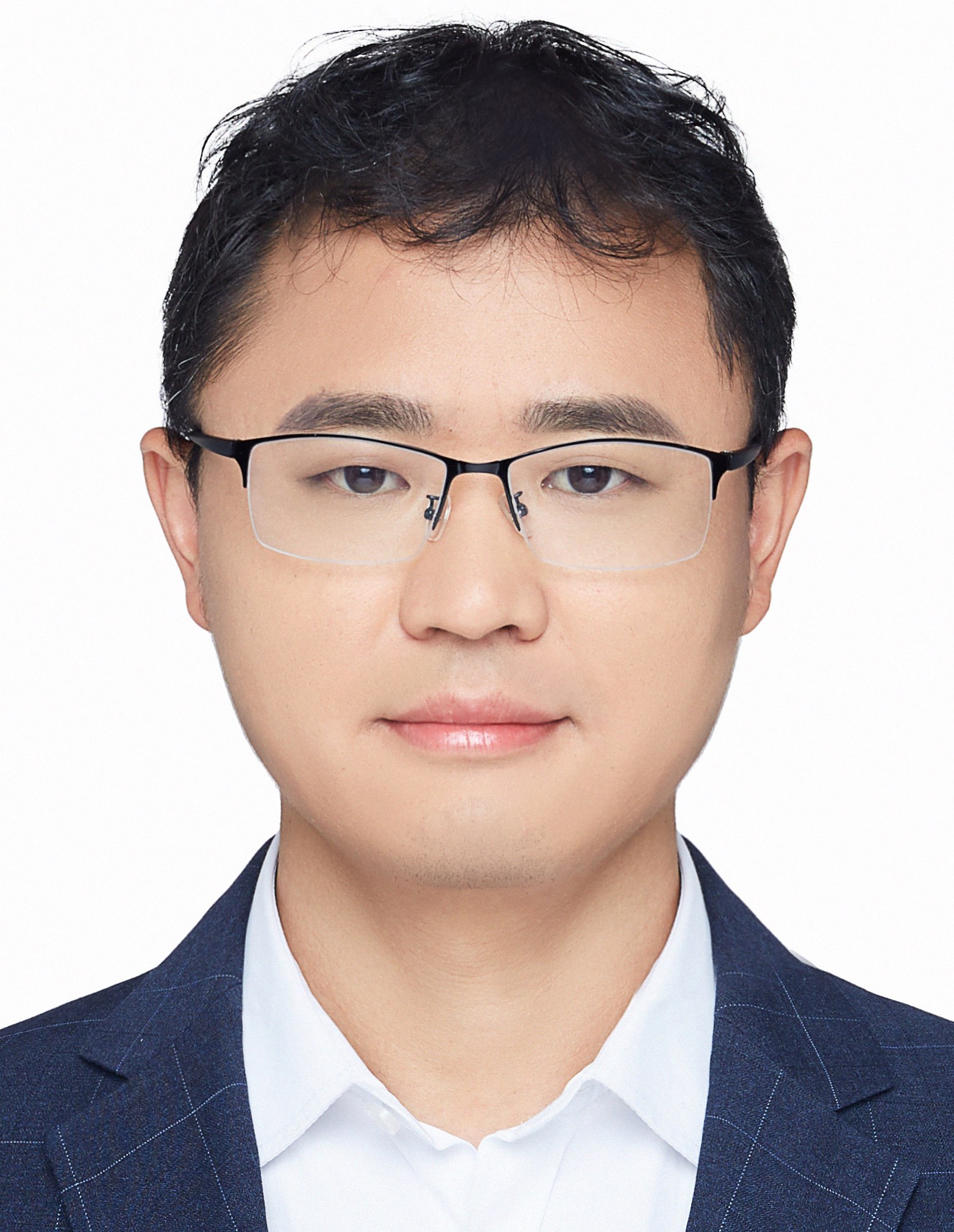}}]{Juyong Zhang}
is a professor in the School of Mathematical Sciences at University of Science and Technology of China. He received the BS degree in Computer Science and Engineering from the University of Science and Technology of China in 2006, and the PhD degree from Nanyang Technological University, Singapore in 2011. His research interests include computer graphics, computer vision, and numerical optimization. He is an associate editor of IEEE Transactions on Multimedia.

\end{IEEEbiography}

\clearpage

\end{document}